\documentclass[10pt,twocolumn,letterpaper]{article}
\usepackage[pagenumbers]{cvpr}      % To produce the REVIEW version
\usepackage[accsupp]{axessibility}% \usepackage[pagenumbers]{cvpr} % To force page numbers, e.g. for an arXiv version
\newcommand{\TODO}[1]{\textbf{\color{red}[TODO: #1]}}
\usepackage{makecell}
\usepackage{pifont}

\newcommand{\statusDone}[1]{#1}

\newcommand{\statusWIP}[1]{#1}

\usepackage{framed}

\usepackage[usestackEOL]{stackengine}

\usepackage{epsfig}
\usepackage{adjustbox} % Extends graphicx and \includegraphics
\usepackage[dvipsnames]{xcolor} % Seems to be loaded by the style sheet

\usepackage{tikz}
\usetikzlibrary{decorations.pathreplacing,calc,arrows,automata}

\graphicspath{{./figures/}}
\DeclareGraphicsExtensions{.pdf,.png,.jpeg,.jpg}

\usepackage{bm} % Bold maths
\usepackage{esint} % More integrals for computer modern math fonts
\usepackage{commath} % Provides \dif upright d

\usepackage{siunitx} % For typesetting values wih units
\usepackage{subcaption} % Recommended, e.g., with the latest ACM template

\usepackage{booktabs} % Better-looking, more formal tables
\usepackage{enumitem} % Better control of lists and their spacing

\usepackage{multirow}
\usepackage{tikz}
\usetikzlibrary{decorations.pathreplacing,calc,arrows,automata}
\newcommand{\tikzmark}[2][-3pt]{\tikz[remember picture, overlay, baseline=-0.5ex]\node[#1](#2){};}

\renewcommand{\mid}{\,\ifnum\currentgrouptype=16 \middle\fi|\,}

\newcommand{\new}[1]{{\color{blue}#1}} % For highlighting revisions
\usepackage{colortbl} % For \cellcolor

\usepackage{boxedminipage} % Creates boxed minipages that do not extend into the margins
\newcommand{\GENEA}{AGG} % Anonymise the GENEA name for the leaderboard
\let\oldmarginpar\marginpar
\renewcommand\marginpar[1]{\-\oldmarginpar[\raggedleft\footnotesize #1]%
{\raggedright\footnotesize #1}}

\usepackage{pdfrender}
\usepackage{pgf-pie} % in your preamble
\usepackage{tikz}
\usepackage{soul}
\usepackage{caption}
\usepackage{placeins}
\makeatletter
\tikzstyle{notestyleraw}=[
  draw=\@todonotes@currentbordercolor,
  fill=\@todonotes@currentbackgroundcolor,
  line width=0.5pt,
  text width=\@todonotes@textwidth-1.6ex-1pt,
  inner sep=0.8ex
]
\makeatother
\usepackage{tikz} % load TikZ first
\usepackage[most]{tcolorbox}
\usepackage{afterpage}
\usepackage{pgfplots}
\pgfplotsset{compat=1.18}
\usepgfplotslibrary{groupplots} % Load the library

\newenvironment{hlbox}
{
\begin{tcolorbox}[
  enhanced,
  rounded corners,
  colback=yellow!10!white,             % very light gray background
  colframe=black!20,          % subtle gray border
  boxrule=0.3pt,              % thin border
  left=3pt, right=3pt, top=2pt, bottom=2pt, % compact padding
  boxsep=1pt,                 % tight inner spacing
  before skip=4pt, after skip=4pt, % compact vertical spacing
  fonttitle=\bfseries,
  coltitle=black,
  colbacktitle=white,
  boxed title style={empty},  % no colored title bar
]
}
{
\end{tcolorbox}
}
\definecolor{cvprblue}{rgb}{0.21,0.49,0.74}
\usepackage[pagebackref,breaklinks,colorlinks,allcolors=cvprblue]{hyperref}

\usepackage[capitalize]{cleveref} % Support for easy cross-referencing
\usepackage{lipsum}
\Crefname{section}{Section}{Sections}
\crefname{section}{Sec.}{Secs.}
\Crefname{table}{Table}{Tables}
\crefname{table}{Table}{Tables}

\def\paperID{18014} % *** Enter the Paper ID here
\def\confName{CVPR}
\def\confYear{2026}

\title{Towards Reliable Human Evaluations in Gesture Generation: Insights from a Community-Driven State-of-the-Art Benchmark}
\author{
Rajmund Nagy\textsuperscript{1},
Hendric Voss\textsuperscript{2},
Thanh Hoang-Minh\textsuperscript{3},
Mihail Tsakov\textsuperscript{4},
Teodor Nikolov\textsuperscript{5},
\\
Zeyi Zhang\textsuperscript{6},
Tenglong Ao\textsuperscript{6},
Sicheng Yang\textsuperscript{7},
Shaoli Huang\textsuperscript{8},
Yongkang Cheng\textsuperscript{8},\\
M. Hamza Mughal\textsuperscript{9},
Rishabh Dabral\textsuperscript{9},
Kiran Chhatre\textsuperscript{1},
Christian Theobalt\textsuperscript{9},
Libin Liu\textsuperscript{6},
Stefan Kopp\textsuperscript{2},
\\
Rachel McDonnell\textsuperscript{10},
Michael Neff\textsuperscript{11},
Taras Kucherenko\textsuperscript{12},
Youngwoo Yoon\textsuperscript{13}\thanks{Co-corresponding authors.},
Gustav Eje Henter\textsuperscript{1,5}\footnotemark[1]
\\[1ex]
\textsuperscript{1}KTH Royal Institute of Technology
\textsuperscript{2}Bielefeld University
\textsuperscript{3}University of Science -- VNUHCM\\
\textsuperscript{4}Independent Researcher
\textsuperscript{5}Motorica AB
\textsuperscript{6}Peking University
\textsuperscript{7}Tsinghua University
\textsuperscript{8}Astribot \\
\textsuperscript{9}Max-Planck Institute for Informatics, SIC
\textsuperscript{10}Trinity College Dublin
\textsuperscript{11}University of California, Davis\\
\textsuperscript{12}SEED -- Electronic Arts
\textsuperscript{13}Electronics and Telecommunications Research Institute (ETRI)
}

\begin{document}
\maketitle

% *** ABSTRACT ***
\begin{abstract}
We review human evaluation practices in automatic, speech-driven 3D gesture generation and find a lack of standardisation and frequent use of flawed experimental setups. This leads to a situation where it is impossible to know how different methods compare, or what the state of the art is.

In order to address common shortcomings of evaluation design, and to standardise future user studies in gesture-generation works, we introduce a detailed human evaluation protocol for the widely-used BEAT2 motion-capture dataset. Using this protocol, we conduct large-scale crowdsourced evaluation to rank six recent gesture-generation models -- each trained by its original authors -- across two key evaluation dimensions: motion realism and speech-gesture alignment. 

Our results show that 1) motion realism has become a saturated evaluation measure on the BEAT2 dataset, with older models performing on par with more recent approaches; 
2) previous findings of high speech-gesture alignment do not hold up under rigorous evaluation, even for specialised models; and 3) the field must adopt disentangled assessments of motion quality and multimodal alignment for accurate benchmarking in order to make progress.

To drive standardisation and enable new evaluation research, we release five hours of synthetic motion from the benchmarked models; over 750 rendered video stimuli from the user studies -- enabling new evaluations without requiring model reimplementation -- alongside our open-source rendering script, and 16,000 pairwise human preference votes collected for our benchmark.

\end{abstract}

% *** BODY TEXT ***
\section{\statusDone{Introduction}}
\label{sec:intro}
% The introduction should essentially not teach the reader anything about the world, but teach them about the paper
Research interest in automatic gesture generation -- the task of animating speaking 3D characters -- has been sharply rising as part of the recent boom in generative and multimodal AI \cite{liu2021speech,nyatsanga_comprehensive_2023,abootorabi2025generative}. However, whilst the latest generative models are being applied to this domain, trustworthy empirical evaluation of communicative non-verbal behaviour remains understudied. Even human evaluations, widely considered the gold standard, may lead to misleading conclusions \cite{bylinskii2023towards, hertzmann2023curse} due to their complexity. Without carefully designed evaluation standards adopted on a community level, published research may give a false sense of progress.

We identify several dimensions of evaluation design in automatic gesture-generation where the lack of standardisation leads to unreliable findings and incomparable studies. To address this, we develop an evaluation protocol on BEAT2 \cite{liu2024emage}, the most widely adopted speech-gesture dataset according to our survey, building on the methodology of the GENEA Challenges \cite{kucherenko2023genea, kucherenko2021large, kucherenko2024evaluating} with improvements to validity and re-usability. 

To validate our protocol% and to understand the state of the art in gesture generation
, we benchmark six recently published gesture-generation models using crowdsourced human evaluation% following the proposed methodology
. We present findings that contradict previously published claims, and release all collected information in the process\footnote{Available on 
 \href{https://genea-workshop.github.io/leaderboard/}{https://genea-workshop.github.io/leaderboard/}.} to enable future research on gesture-generation evaluation. To summarise, our contributions are:
\begin{enumerate}
    \item A critical \textbf{review} of evaluation practices (\cref{sec:limitations-survey});
    \item A new \textbf{evaluation protocol} for the widely-used BEAT2 dataset, rooted in prior large-scale evaluations (\cref{sec:protocol});
    \item A new \textbf{benchmark} of six recent models to illuminate the state of the art (\cref{sec:experiments}). 
\end{enumerate}

\section{\statusDone{Background}}
\label{sec:background}
\subsection{\statusDone{Automatic Gesture Generation}}
\label{ssec:automatic-gesturing}
We define speech-driven 3D gesture generation as the problem of mapping from a time-aligned input speech sequence $s$ (e.g., an audio waveform and/or a text transcription) with potential additional conditioning $c$ (e.g., an emotion label), to a 3D human body-motion sequence output $x$ (e.g., a series of joint angles defining human poses). Whilst early deep-learning-based systems for gesture generation treated gesture generation as a regression problem \cite{takeuchi2017speech,hasegawa2018evaluation,ferstl2018trinity,yoon2019robots,kucherenko2019analyzing,kucherenko2021moving,kucherenko2020gesticulator,yoon2020speech}, recent approaches use probabilistic, generative models like normalising flows \cite{alexanderson2020style}, VAEs \cite{ghorbani2023zeroeggs}, VQ-VAEs \cite{yazdian2022gesture2vec,voss2023aq}, discrete autoregressive transformer models \cite{yang2023qpgesture}, or diffusion models \cite{zhu2023taming,ao2023gesturediffuclip, alexanderson2023listen, yang2023diffusestylegesture,deichler2023diffusion, mughal2024convofusion,chhatre2024amuse, ng2024audio2photoreal}. 

\subsection{\statusDone{Evaluations in Automatic Gesture Generation}}
\label{ssec:evaluation}
Both automatic (``objective'') as well as human (``subjective'') evaluation methodologies are widely used in 3D gesture generation \cite{nyatsanga_comprehensive_2023}. While automatic metrics like FGD \cite{yoon2020speech} or beat consistency \cite{liu2022learning} can be useful to guide model development, they have well-known limitations and unclear relationships with human perception \cite{yang2023qpgesture,kucherenko2024evaluating,Crnek2025Advancing,tseng2023edge, dabral2022mofusion, haque2025wild, yang2024probabilistic}. Therefore, final results are almost universally validated by analysing responses given by human evaluators.

\begin{hlbox}
Human assessment must thus be considered the gold-standard evaluation methodology in gesture generation, and is the key focus of this paper.
\end{hlbox} 

\subsubsection{Standardisation Needs}
\label{ssec:background-standardisation-needs}
Despite being considered the gold standard, there is little information available on the ecological validity of human-evaluation practices in gesture generation. A 2021 review by \citet{wolfert2021review} %of 22 gesture-generation methods predominantly published in 2018 or earlier 
found a large diversity in both automatic and human evaluation methods, with poor reporting practices on participant characteristics and evaluation design, concluding that the field would ``\emph{benefit from more experimental rigour and a shared methodology for conducting systematic evaluation}''. 

The GENEA Challenges \cite{kucherenko2021large, yoon2022genea, kucherenko2023genea, kucherenko2024evaluating} were launched in 2020 to address this problem by evaluating sets of gesture-generation models under standardised conditions. The challenges are a series of community-driven evaluations of automatic gesture-generation models. Each challenge collected between 5 and 12 model submissions trained on a selected dataset, which were then subjected to large-scale crowdsourced human evaluation by the organisers. Challenge findings emphasise the importance of data filtering, removing artifacts, high-quality visualisations, and disentangled evaluation (as discussed in \cref{ssec:importanceofmismatching}), and provide strong evidence for the importance of standardised human evaluation. However, whether the GENEA Challenges address the problem of missing standardisation is an open question. There has been no assessment of whether their methodology is adopted by the community, and their results form isolated user studies rather than a continuously growing benchmark. Furthermore, the GENEA Challenges have seldom included systems from major computer vision or machine-learning conferences like CVPR or SIGGRAPH, and therefore their results may not reflect the state of the art. 

\section{Key Limitations of Current Human-Evaluation Practices}
\label{sec:limitations-survey}
% Without standardised benchmarking practices, the state of the art can only be assessed by systematically reviewing independent evaluation results of published models. Are these evaluations reliable, or do inconsistencies in evaluation protocols, participant sampling, and reporting practices undermine their validity? Can we meaningfully assess the state of the art by comparing results from competing models when datasets, visualisation methods, and human-evaluation designs differ so widely across studies, or if such comparisons are fundamentally flawed?

%This is especially important as it is currently unclear if the evaluation results reported in recently published 3D gesture-generation papers are reliable, or if inconsistencies in evaluation protocols, participant sampling, and reporting practices undermine their validity. We therefore also need to ask the question, if we can meaningfully assess the state of the art by comparing results from competing models when datasets, visualisation methods, and human-evaluation designs differ so widely across studies, or if such comparisons are fundamentally flawed?

We perform a critical assessment of human evaluation practices in recently published gesture-generation research, aiming to answer the following questions:
\begin{enumerate}
    \item Are evaluation results \emph{reliable}, in that they measure what they purport to measure?
    \item Are evaluation results \emph{comparable} between different publications? Can we assess the state of the art from independent evaluations? 
\end{enumerate}
We review 26 recent publications on co-speech gesture-generation methods from selected computer vision and graphics conferences (CVPR, ICCV, ECCV, SIGGRAPH, and SIGGRAPH Asia) from 2023 onwards, using the search terms ``gesture'', ``co-speech'', ``speech'', and ``motion'' in publication titles, then filtering down the results to models whose outputs include 3D body gesture. The list of papers is presented on \cref{tab:recent_models}. 

% In the second-to-last column, we note the worrying trend of recent publications not comparing against the reference human gestures (from motion capture). In the last column, we show all direct comparisons amongst the surveyed works driven by human evaluations (cf. \cref{ssec:lack-of-comparisons}): each \textcolor{ACMOrange}{\texttt{\textbf{M}}} entry is indicates that the model in that row was compared to the baseline model \textcolor{ACMDarkBlue}{\texttt{\textbf{B}}} in that column. Overall, the lack of direct comparisons against human motion and strong baseline models makes it challenging to assess the progress made by most publications.
\newcommand{\upbar}{\tikz[overlay] \draw (0,1em)--(0,0em);}
\newcommand{\downbar}{\tikz[overlay] \draw (0,.5em)--(0,-1em);}

\definecolor{lightgreen}{rgb}{0.7,1,0.7}
\definecolor{lightred}{rgb}{1,0.7,0.7}
\definecolor{lightyellow}{rgb}{1,1,0.6}
\newcommand{\colorparticipants}[1]{%
\ifnum#1>100\cellcolor{lightgreen}#1%
\else\ifnum#1>50\cellcolor{lightyellow}#1%
\else\cellcolor{lightred}#1%
\fi\fi}
\definecolor{color_Na}{rgb}{0.106, 0.620, 0.467}
\definecolor{color_Re}{rgb}{0.459, 0.439, 0.702}
\definecolor{color_Hu}{rgb}{0.400, 0.651, 0.118}
\definecolor{color_Sm}{rgb}{0.651, 0.463, 0.114}
\definecolor{color_Pref}{rgb}{0.106, 0.620, 0.467}
\definecolor{color_Rh}{rgb}{0.851, 0.373, 0.008}
\definecolor{color_Sem}{rgb}{0.906, 0.161, 0.541}
\definecolor{color_Gen}{rgb}{0.902, 0.671, 0.008}
\definecolor{color_Em}{rgb}{0.400, 0.400, 0.400}
\definecolor{color_St}{rgb}{0.851, 0.373, 0.008}

\definecolor{ACMGreen}{RGB}{0, 150, 0}
\definecolor{ACMRed}{RGB}{220, 0, 0}
\definecolor{ACMDarkBlue}{RGB}{0, 51, 102}
\definecolor{ACMOrange}{RGB}{255, 102, 0}

\newcounter{brace}
\setcounter{brace}{0}
\newcommand{\drawbrace}[4]{%
  \refstepcounter{brace}
  \tikz[remember picture, overlay]%
    \node at ($(#2.center)+(#3,0)$) {\textcolor{ACMDarkBlue}{\texttt{\textbf{B}}}};%
  \tikz[remember picture, overlay]%
    %\draw[latex-latex,dotted] ($(#1.center)+(#3,-#4)$) -- ($(#2.center)+(#3,#4)$);%
    \draw[dotted] ($(#1.center)+(#3,-#4)$) -- ($(#2.center)+(#3,#4)$);%
  \tikz[remember picture, overlay]%
    \node at ($(#1.center)+(#3,0)$) {\textcolor{ACMOrange}{\texttt{\textbf{M}}}};%
}

\begin{table*}[htb!]
\centering

\caption{Overview of human evaluation practices in 3D gesture-generation research published at SIGGRAPH, SIGGRAPH Asia, and leading computer-vision venues between 2023--2025, as described in \cref{sec:limitations-survey}. The table uncovers the fragmented state of human evaluation, with inconsistent study designs for related tasks (\emph{Modelling Goal} column), and a critically low degree of direct comparisons between top models (last column). Abbreviations: SG=SIGGRAPH; 
\textbf{\textcolor{color_Na}{Na}}=Naturalness;
\textbf{\textcolor{color_Re}{Re}}=Realism, Plausability or Believability;
\textbf{\textcolor{color_Hu}{Hu}}=Human-likeness;
\textbf{\textcolor{color_Sm}{Sm}}=Smoothness;
\textbf{\textcolor{color_Pref}{Pref}}=Preference;
\textbf{\textcolor{color_Rh}{Rh}}=Rhythmic;
\textbf{\textcolor{color_Sem}{Sem}}=Semantic;%\hphanltom{Abbreviations:}
\textbf{\textcolor{color_Gen}{Gen}}=General;
\textbf{\textcolor{color_Em}{Em}}=Emotion;
\textbf{\textcolor{color_St}{St}}=Style;
\textcolor{ACMDarkBlue}{\texttt{\textbf{B}}}=Present in direct comparison as baseline; and
\textcolor{ACMOrange}{\texttt{\textbf{M}}}=Present in direct comparison as main model.}
\label{tab:recent_models}

\begin{tabular}{@{}llllllcl@{}}
\toprule

\multirow{2}{*}{Year} & \multirow{2}{*}{Venue} & \multirow{2}{*}{Model} & \multirow{2}{*}{Training dataset} &
\multicolumn{2}{c}{Modelling Goal} & \multicolumn{2}{c}{Directly compared to...} \\
\cmidrule{5-6}
\cmidrule{7-8}
&&&& Quality & Alignment & Mocap & A model in the survey \\
\midrule

2023 & CVPR  & DiffGesture \cite{zhu2023taming} & TED \cite{yoon2019robots}, TED-Expr. \cite{zhu2023taming} & \textbf{\textcolor{color_Na}{Na}}, \textbf{\textcolor{color_Sm}{Sm}} & \textbf{\textcolor{color_Rh}{Rh}} & \textcolor{ACMGreen}{\ding{52}} &  \tikzmark[xshift=0em]{diffgesture}\hphantom{A model in the survey}\\
\downbar & \downbar & QPGesture \cite{yang2023qpgesture} & BEAT \cite{liu2022beat}& \textbf{\textcolor{color_Hu}{Hu}} & \textbf{\textcolor{color_Gen}{Gen}} & \textcolor{ACMGreen}{\ding{52}} &  \tikzmark[xshift=0em]{qpgesture}\hphantom{A model in the survey}\\
\downbar & \downbar & RACER \cite{sun2023co} & Trinity \cite{ferstl2018trinity}, own & \textbf{\textcolor{color_Re}{Re}} & \textbf{\textcolor{color_Gen}{Gen}}, \textbf{\textcolor{color_Sem}{Sem}} & \textcolor{ACMRed}{\ding{55}} &  \tikzmark[xshift=0em]{racer}\hphantom{A model in the survey}\\
\downbar & \upbar & TalkSHOW \cite{yi2023generating} & SHOW \cite{yi2023generating} & \textcolor{ACMRed}{\ding{55}} & \textbf{\textcolor{color_Gen}{Gen}} & \textcolor{ACMGreen}{\ding{52}} &  \tikzmark[xshift=0em]{talkshow}\hphantom{A model in the survey}\\
\downbar & SG.  & Bodyformer \cite{pang2023bodyformer} & Trinity \cite{ferstl2018trinity}, TWH \cite{lee2019talking} & \textbf{\textcolor{color_Hu}{Hu}} & \textbf{\textcolor{color_Gen}{Gen}} & \textcolor{ACMGreen}{\ding{52}} &  \tikzmark[xshift=0em]{bodyformer}\hphantom{A model in the survey}\\
\downbar & \downbar & GestureDiffuCLIP \cite{ao2023gesturediffuclip} & BEAT \cite{liu2022beat}, ZEGGS \cite{ghorbani2023zeroeggs} & \textbf{\textcolor{color_Hu}{Hu}} & \textbf{\textcolor{color_Sem}{Sem}}, \textbf{\textcolor{color_St}{St}} & \textcolor{ACMGreen}{\ding{52}} &  \tikzmark[xshift=0em]{gesturediffuclip}\hphantom{A model in the survey}\\
\downbar & \upbar & LDA \cite{alexanderson2023listen} & Trinity \cite{ferstl2018trinity}, ZEGGS \cite{ghorbani2023zeroeggs} & \textbf{\textcolor{color_Pref}{Pref}} & \textbf{\textcolor{color_St}{St}} & \textcolor{ACMGreen}{\ding{52}} &  \tikzmark[xshift=0em]{lda}\hphantom{A model in the survey}\\
\downbar & ICCV  & C-DiffGAN \cite{ahuja2023continual} & PATS \cite{ahuja2020no} & \textbf{\textcolor{color_Na}{Na}} & \textbf{\textcolor{color_Sem}{Sem}}, \textbf{\textcolor{color_Rh}{Rh}}, \textbf{\textcolor{color_St}{St}} & \textcolor{ACMGreen}{\ding{52}} &  \tikzmark[xshift=0em]{c-diffgan}\hphantom{A model in the survey}\\
\upbar & \upbar & LivelySpeaker \cite{zhi2023livelyspeaker} & BEAT \cite{liu2022beat}, TED \cite{yoon2019robots} & \textbf{\textcolor{color_Na}{Na}}, \textbf{\textcolor{color_Sm}{Sm}} & \textbf{\textcolor{color_Sem}{Sem}} & \textcolor{ACMRed}{\ding{55}} &  \tikzmark[xshift=0em]{livelyspeaker}\hphantom{A model in the survey}\\

\midrule

2024 & CVPR  & AMUSE \cite{chhatre2024amuse} & BEAT2 \cite{liu2024emage} & \textcolor{ACMRed}{\ding{55}} & \textbf{\textcolor{color_Rh}{Rh}}, \textbf{\textcolor{color_Em}{Em}} & \textcolor{ACMGreen}{\ding{52}} &  \tikzmark[xshift=0em]{amuse}\hphantom{A model in the survey}\\
\downbar & \downbar & Audio2Photoreal \cite{ng2024audio2photoreal} & own & \textbf{\textcolor{color_Re}{Re}} & \textcolor{ACMRed}{\ding{55}} & \textcolor{ACMGreen}{\ding{52}} &  \tikzmark[xshift=0em]{audio2photoreal}\hphantom{A model in the survey}\\
\downbar & \downbar & ConvoFusion \cite{mughal2024convofusion} & DnD \cite{mughal2024convofusion} & \textbf{\textcolor{color_Na}{Na}} & \textbf{\textcolor{color_Gen}{Gen}}, \textbf{\textcolor{color_Sem}{Sem}} & \textcolor{ACMGreen}{\ding{52}} &  \tikzmark[xshift=0em]{convofusion}\hphantom{A model in the survey}\\
\downbar & \downbar & DiffSHEG \cite{chen2024diffsheg} & BEAT \cite{liu2022beat}, SHOW \cite{yi2023generating}& \textbf{\textcolor{color_Re}{Re}} & \textbf{\textcolor{color_Rh}{Rh}} & \textcolor{ACMRed}{\ding{55}} &  \tikzmark[xshift=0em]{diffsheg}\hphantom{A model in the survey}\\
\downbar & \downbar & EMAGE \cite{liu2024emage} & BEAT2 \cite{liu2024emage} & \textbf{\textcolor{color_Re}{Re}} & \textcolor{ACMRed}{\ding{55}} & \textcolor{ACMRed}{\ding{55}} &  \tikzmark[xshift=0em]{emage}\hphantom{A model in the survey}\\
\downbar & \downbar & EmoTransition \cite{qi2024emotransition} & own & \textbf{\textcolor{color_Na}{Na}}, \textbf{\textcolor{color_Sm}{Sm}} & \textcolor{ACMRed}{\ding{55}} & \textcolor{ACMRed}{\ding{55}} &  \tikzmark[xshift=0em]{emotransition}\hphantom{A model in the survey}\\
\downbar & \upbar & ProbTalk \cite{liu2024towards_probtalk} & SHOW \cite{yi2023generating} & \textcolor{ACMRed}{\ding{55}} & \textcolor{ACMRed}{\ding{55}} & \textcolor{ACMRed}{\ding{55}} &  \tikzmark[xshift=0em]{probtalk}\hphantom{A model in the survey}\\
\downbar & SG.  & Sem. Gest. \cite{zhang2024semantic} & BEAT \cite{liu2022beat}, ZEGGS \cite{ghorbani2023zeroeggs}, own & \textbf{\textcolor{color_Hu}{Hu}} & \textbf{\textcolor{color_Sem}{Sem}}, \textbf{\textcolor{color_Rh}{Rh}} & \textcolor{ACMGreen}{\ding{52}} &  \tikzmark[xshift=0em]{semgest}\hphantom{A model in the survey}\\
\upbar & SG. Asia  & SIGGesture \cite{cheng2024siggesture} & BEAT \cite{liu2022beat}& \textbf{\textcolor{color_Na}{Na}} & \textbf{\textcolor{color_Sem}{Sem}}, \textbf{\textcolor{color_Rh}{Rh}} & \textcolor{ACMRed}{\ding{55}} &  \tikzmark[xshift=0em]{siggesture}\hphantom{A model in the survey}\\

\midrule

2025 & CVPR  & HOP \cite{cheng2025hop} & TED \cite{yoon2019robots}, TED-Expr. \cite{zhu2023taming} & \textbf{\textcolor{color_Na}{Na}}, \textbf{\textcolor{color_Sm}{Sm}} & \textbf{\textcolor{color_Sem}{Sem}}, \textbf{\textcolor{color_Rh}{Rh}} & \textcolor{ACMGreen}{\ding{52}} &  \tikzmark[xshift=0em]{hop}\hphantom{A model in the survey}\\
\downbar & \downbar & LOM \cite{chen2024language} & BEAT2 \cite{liu2024emage} & \textcolor{ACMRed}{\ding{55}} & \textcolor{ACMRed}{\ding{55}} & \textcolor{ACMRed}{\ding{55}} &  \tikzmark[xshift=0em]{lom}\hphantom{A model in the survey}\\
\downbar & \upbar & RAG-Gesture \cite{mughal2024raggesture} & BEAT2 \cite{liu2024emage} & \textbf{\textcolor{color_Na}{Na}} & \textbf{\textcolor{color_Sem}{Sem}} & \textcolor{ACMGreen}{\ding{52}} &  \tikzmark[xshift=0em]{rag-gesture}\hphantom{A model in the survey}\\
\downbar & SG.  & MeCo \cite{chen2025meco} & BEAT2 \cite{liu2024emage}, ZEGGS \cite{ghorbani2023zeroeggs} & \textbf{\textcolor{color_Hu}{Hu}} & \textbf{\textcolor{color_Gen}{Gen}} & \textcolor{ACMRed}{\ding{55}} &  \tikzmark[xshift=0em]{meco}\hphantom{A model in the survey}\\
\downbar & ICCV  & GestureHydra \cite{yang2025GestureHYDRA} & SHOW \cite{yi2023generating}, own & \textbf{\textcolor{color_Na}{Na}} & \textbf{\textcolor{color_Sem}{Sem}} & \textcolor{ACMRed}{\ding{55}} &  \tikzmark[xshift=0em]{gesturehydra}\hphantom{A model in the survey}\\
\downbar & \downbar & GestureLSM \cite{liu2025gesturelsmlatentshortcutbased} & BEAT2 \cite{liu2024emage} & \textbf{\textcolor{color_Sm}{Sm}}, \textbf{\textcolor{color_Re}{Re}} & \textbf{\textcolor{color_Rh}{Rh}} & \textcolor{ACMRed}{\ding{55}} &  \tikzmark[xshift=0em]{gesturelsm}\hphantom{A model in the survey}\\
\downbar & \downbar & SemGes \cite{liu2025semges} & BEAT \cite{liu2022beat}, TED-Expr. \cite{zhu2023taming} & \textbf{\textcolor{color_Na}{Na}} & \textbf{\textcolor{color_Sem}{Sem}}, \textbf{\textcolor{color_Rh}{Rh}} & \textcolor{ACMGreen}{\ding{52}} &  \tikzmark[xshift=0em]{semges}\hphantom{A model in the survey}\\
\upbar & \upbar & SemTalk \cite{zhang2024semtalk} & BEAT2 \cite{liu2024emage}, SHOW \cite{yi2023generating} & \textbf{\textcolor{color_Re}{Re}} & \textbf{\textcolor{color_Sem}{Sem}}, \textbf{\textcolor{color_Rh}{Rh}} & \textcolor{ACMRed}{\ding{55}} &  \tikzmark[xshift=0em]{semtalk}\hphantom{A model in the survey}\\

\bottomrule
\iffalse
\multicolumn{8}{l}{\scriptsize \parbox[t]{0.8\linewidth}{Abbreviations: 
\textbf{\textcolor{color_Na}{Na}}=Naturalness;
\textbf{\textcolor{color_Re}{Re}}=Realism, Plausability or Believability;
\textbf{\textcolor{color_Hu}{Hu}}=Human-likeness;
\textbf{\textcolor{color_Sm}{Sm}}=Smoothness;
\textbf{\textcolor{color_Pref}{Pref}}=Preference;
\textbf{\textcolor{color_Rh}{Rh}}=Rhythmic;
\textbf{\textcolor{color_Sem}{Sem}}=Semantic;%\hphantom{Abbreviations:}
\textbf{\textcolor{color_Gen}{Gen}}=General;
\textbf{\textcolor{color_Em}{Em}}=Emotion;
\textbf{\textcolor{color_St}{St}}=Style;
\textcolor{ACMDarkBlue}{\texttt{\textbf{B}}}=Present in direct comparison as baseline; and
\textcolor{ACMOrange}{\texttt{\textbf{M}}}=Present in direct comparison as main model.
}}
\fi
\end{tabular}
\end{table*}

\subsection{Entanglement Between Evaluation Dimensions}
\label{ssec:review-finding-disentanglement}
We find a standard practice among reviewed works to conduct human evaluations across two dimensions\footnote{See \cref*{tab:recent_models} for details.}:
\begin{enumerate}
\item \emph{Motion realism}, i.e., whether gestures look sufficiently natural and visually convincing to a human observer. 
\item \emph{Multimodal alignment} between the output motion and the inputs (most commonly, speech); also known as multimodal grounding \cite{nyatsanga_comprehensive_2023} and appropriateness \cite{kucherenko2021large, kucherenko2024evaluating, kucherenko2023genea}.
\end{enumerate}
We find that evaluations of motion realism and multimodal alignment are generally conducted using the same user-study setup -- where participants compare videos of speaking characters animated by different models -- with only the evaluation question changed in between. We will refer to this as the \emph{naive approach}. Despite its widespread usage, there is evidence that the naive approach fails to separate motion realism and multimodal alignment. 

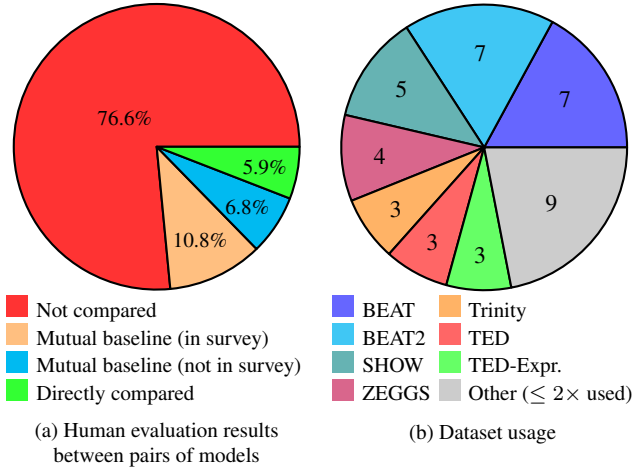
\begin{figure}[t]
  \centering
  % Tighter subcaption layout
  \captionsetup[subfigure]{justification=centering,singlelinecheck=false,margin=0pt,skip=2pt}
  \begin{subfigure}[t]{0.48\linewidth}
    \centering
    \begin{tikzpicture}
      \pie[
        text=,
        font=\footnotesize,
        radius=1.9,
        sum=auto,
        after number=\%,
        color={red!80, orange!50, cyan!80, green!80},
      ]{
        76.6/Not compared,
        10.8/Shared baseline (not in survey),
        6.8/Shared baseline (in survey),
        5.9/Directly compared
      }
    \end{tikzpicture}
    \vspace{0.25em}
    {\footnotesize
      \setlength{\tabcolsep}{0.2em}\renewcommand{\arraystretch}{0.9}%
      \begin{tabular}{@{}ll@{}}
        \textcolor{red!80}{\rule{1em}{1em}} & Not compared \\
        \textcolor{orange!50}{\rule{1em}{1em}} & Mutual baseline (in survey) \\
        \textcolor{cyan!80}{\rule{1em}{1em}} & Mutual baseline (not in survey) \\
        \textcolor{green!80}{\rule{1em}{1em}} & Directly compared \\
      \end{tabular}
    }
    \subcaption{Human evaluation results between pairs of models}\label{fig:review-direct-comparisons}
  \end{subfigure}
  \hfill
  \begin{subfigure}[t]{0.48\linewidth}
    \centering
    \begin{tikzpicture}
      % The pie chart data is updated as per your request
      \pie[
        text=,
        font=\small,
        radius=1.9,
        sum=auto,
        % A new color (green!60) has been added for the new data category
        color={blue!60, cyan!60, teal!60, purple!60, orange!60, red!60, green!60, gray!40}
      ]{
        7/BEAT,
        7/BEAT2,
        5/SHOW,
        4/ZEGGS,
        3/Trinity,
        3/TED,
        3/TED-Expressive,
        9/Other
      }
    \end{tikzpicture}
    \vspace{0.25em}
    {\footnotesize
      \setlength{\tabcolsep}{0.2em}\renewcommand{\arraystretch}{0.9}%
      \begin{tabular}{@{}llll@{}}
        \textcolor{blue!60}{\rule{1em}{1em}} & BEAT & \textcolor{orange!60}{\rule{1em}{1em}} & Trinity \\
        \textcolor{cyan!60}{\rule{1em}{1em}} & BEAT2 & \textcolor{red!60}{\rule{1em}{1em}} & TED \\
        \textcolor{teal!60}{\rule{1em}{1em}} & SHOW & \textcolor{green!60}{\rule{1em}{1em}} & TED-Expr. \\
        \textcolor{purple!60}{\rule{1em}{1em}} & ZEGGS & \textcolor{gray!40}{\rule{1em}{1em}} & Other ($\leq2\times$ used) \\
      \end{tabular}
    }
    \subcaption{Dataset usage}\label{fig:review-dataset-usage}
  \end{subfigure}
  \caption{Direct comparisons are exceedingly rare between state-of-the-art gesture generation models, partially due to spread out dataset usage. Data from 26 surveyed models published at CVPR, ICCV, ECCV, SIGGRAPH, and SIGGRAPH Asia between 2023 and 2025.}
  \label{fig:review-comparisons-datasets}
\end{figure}

\subsubsection{The Importance of Mismatching Evaluations}
\label{ssec:importanceofmismatching}
In 2020, the first GENEA Challenge \cite{kucherenko2021large} used the naive approach to evaluate motion realism and speech-gesture alignment. Surprisingly, models with visually appealing motion received relatively high ratings for speech-gesture alignment even when they did not depend on the speech. In fact, \citet{kucherenko2024evaluating} re-analysed the data from the GENEA Challenge 2020, and found a Pearson correlation of over 0.5 between the mean rated alignment of a given motion segment and its rated motion realism. This is despite the fact that their crowdsourced test takers were instructed not to pay attention to human-likeness when assessing alignment. Therefore, there is strong evidence that:

\begin{hlbox}
Direct evaluations of multimodal alignment -- e.g., when conditioning on speech, emotion, style, or semantic information -- are ineffective unless one controls for the significant confounding effect of motion realism. 
\end{hlbox}
\drawbrace{emotransition}{diffgesture}{1em}{0em}
\drawbrace{siggesture}{qpgesture}{2em}{0em}
\drawbrace{amuse}{talkshow}{3em}{0em}
\drawbrace{diffsheg}{talkshow}{3em}{0em}
\drawbrace{emage}{talkshow}{3em}{0em}
\drawbrace{siggesture}{talkshow}{3em}{0em}
\drawbrace{gesturehydra}{talkshow}{3em}{0em}
\drawbrace{semgest}{gesturediffuclip}{4em}{0em}
\drawbrace{audio2photoreal}{lda}{5em}{0em}
\drawbrace{diffsheg}{lda}{5em}{0em}
\drawbrace{gesturelsm}{diffsheg}{6em}{0em}
\drawbrace{semges}{diffsheg}{6em}{0em}
\drawbrace{siggesture}{emage}{7em}{0em}
\drawbrace{rag-gesture}{emage}{7em}{0em}
\drawbrace{meco}{emage}{7em}{0em}
\drawbrace{gesturelsm}{emage}{7em}{0em}
\drawbrace{semtalk}{emage}{7em}{0em}
\drawbrace{gesturehydra}{probtalk}{8em}{0em}
\drawbrace{gesturelsm}{probtalk}{8em}{0em}

To combat the above problem, later GENEA Challenges \cite{yoon2022genea, kucherenko2023genea, kucherenko2024evaluating} proposed a new evaluation paradigm leveraging the principle of \emph{mismatching} \cite{ennis2010seeing,rebol2021passing,jonell2020let,yoon2022genea}. In this paradigm, so-called \emph{mismatched stimuli} are created by swapping the data for a given modality to an unrelated sequence -- for example, the audio is replaced by a different sentence, or the movements of the conversational partner are replaced by that of another character. Human evaluators, unaware of the mismatching manipulation, are then asked to indicate their preference between the original (matched) stimulus and its mismatched counterpart, based on which one seems more coherent (e.g., in terms of hand movement matching the speech). This setup effectively isolates the strong confounding effect of intrinsic motion quality from multimodal alignment, as the motion in both stimuli is always generated by the same system. Indeed, the GENEA Challenges 2022 and 2023 confirm the efficacy of this paradigm, reporting distinct model rankings for motion quality versus multimodal alignment \cite{yoon2022genea,kucherenko2023genea} and substantially reduced Pearson correlation between the two \cite{kucherenko2024evaluating}.

% A key result in the evaluation4 of gesture generation is that under naive, direct evaluations, motion quality substantially confounds the perception of gesture-speech appropriateness (\cref{ssec:importanceofmismatching}). To resolve this, the GENEA Challenges \cite{yoon2022genea, kucherenko2023genea} proposed a \emph{mismatching} methodology as an effective resolution. However:

\begin{hlbox}
However, our review finds that \emph{none} of the multimodal-alignment evaluations in the 26 reviewed works adopt the mismatching methodology of the GENEA Challenges, nor do they employ other strategies to control for the confounding effect of motion quality. 
\end{hlbox}

This methodological gap is critical since it allows models that merely produce smooth, high-quality motion -- regardless of how well they align with the speech content -- to receive inflated multimodal alignment scores. Consequently, recent findings where speech-gesture alignment in model outputs closely approximates or surpasses the scores of human mocap (e.g., \cite{yang2023qpgesture, yi2023generating, zhu2023taming, mughal2024raggesture}) are not reliable indicators of the generated gestures being on par with the dataset. (We provide strong empirical support for this claim in \cref{ssec:our-alignment-results}.)
% In particular, we are able to evaluate the appropriateness of \citet{mughal2024raggesture} using a best-practices mismatching user study in \cref{ssec:appropriateness-results}, with a comparative analysis between their and our findings provided in \cref{ssec:model-diff}.

\subsection{Lack of Direct Comparisons} \label{ssec:lack-of-comparisons}
We investigate what kind of evaluation results are available when comparing the 26 models included in our survey. 
%For example, the strongest form of evidence for picking between two models comes from a direct comparison. If that is not available, it might still be possible to rank two models if one of them shows larger relative improvement compared to a mutual baseline model. However, if two publications consider non-overlapping sets of models in their evaluation, it becomes infeasible to infer which model works better without running a new evaluation. 
Unfortunately, we uncover a remarkably low degree of direct comparisons between surveyed models, available only for 19 out of 325 possible model pairs. As visualised in \cref{fig:review-comparisons-datasets}, this amounts to \textbf{less than 6\% of evaluation coverage of all possible pairs} of models. % in \cref{tab:recent_models}.
Furthermore, almost three-quarters of the possible pairings between surveyed publications do not have relevant evaluation results. 
%The right-hand pie chart highlights one reason for this fragmentation: the research community is divided across multiple datasets, with no single benchmark dominating usage. While BEAT and BEAT2 are the most common, a significant proportion of studies use smaller or proprietary datasets, which impedes reproducibility and consistent comparison even further.

It is unreasonable to expect every model to be compared to every other model in our review: baselines may be chosen from other conferences, and it is difficult to compare to models trained on other datasets or lacking publicly available implementation. At the same time, the lack of direct comparisons between competing models is glaring, since 10 of the 19 comparisons are against weak baseline models introduced with a dataset \cite{yi2023generating, liu2024emage}. Human evaluations can depend strongly on which other systems are present in the evaluation, and the range of performance they exhibit \cite{cooper2023investingating}. For this reason, the lack of direct comparisons and relevant baseline choices between leading models, as uncovered by our review, pose a serious practical challenge for determining the state of the art.
\begin{figure}[t]
  \centering
  \captionsetup[subfigure]{justification=centering,singlelinecheck=false,margin=0pt,skip=0pt}

  % --- Left subplot: Quality descriptors ---
  \begin{subfigure}[t]{0.48\linewidth}
    \centering
    \begin{tikzpicture}
      \begin{axis}[
        xbar,
        xmin=0,
        xmax=10,
        width=4cm,
        height=3cm,
        bar width=6pt,
        axis y line*=left,
        axis x line*=bottom,
        enlarge y limits=0.2,
        symbolic y coords={Natural,Humanlike,Smooth,Diverse,Realistic},
        ytick=data,
        yticklabels={Natural,Humanlike,Smooth,Diverse,Realistic},
        yticklabel style={font=\footnotesize, align=right},
        tick label style={font=\footnotesize, color=black}
    ]
        \addplot+[draw=none, fill=blue!70] coordinates {
          (10,Natural)
          (5,Humanlike)
          (5,Smooth)
          (5,Diverse)
          (4,Realistic)
        };
      \end{axis}
    \end{tikzpicture}
    \subcaption{Quality descriptors}
    \label{subfig:motion_quality_tags}
  \end{subfigure}
  \hfill
  % --- Right subplot: Forms of alignment ---
  \begin{subfigure}[t]{0.49\linewidth}
    \centering
    \begin{tikzpicture}
      \begin{axis}[
        xbar,
        xmin=0,
        xmax=12,
        width=4cm,
        height=3cm,
        bar width=6pt,
        axis y line*=left,
        axis x line*=bottom,
        enlarge y limits=0.2,
        symbolic y coords={Semantics,Rhythm,Unspecified,Style},
        ytick=data,
        yticklabels={Semantics,Rhythm,Unspecified,Style},
        yticklabel style={font=\footnotesize, align=right},
        tick label style={font=\footnotesize, color=black}
      ]
        \addplot+[draw=none, fill=orange!80] coordinates {
          (12,Semantics)
          (10,Rhythm)
          (6,Unspecified)
          (3,Style)
        };
      \end{axis}
    \end{tikzpicture}
    \subcaption{Forms of alignment}
    \label{subfig:grounding_factors}
  \end{subfigure}

  \caption{Distribution of aspects measured in user studies evaluating motion realism (left) and multimodal alignment (right).}
  \label{fig:review-question-formulations}
\end{figure}
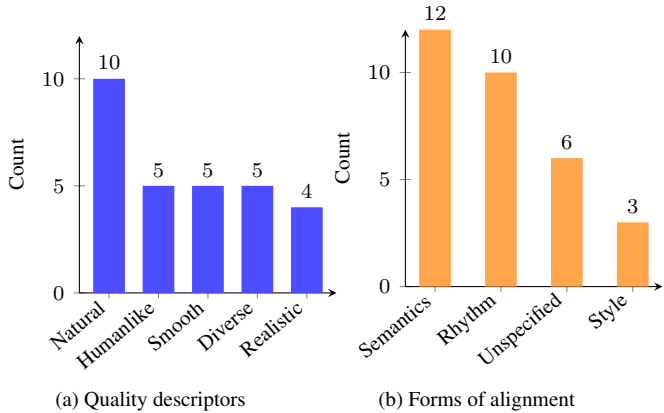

\subsection{Inconsistent, Incompatible Evaluation Designs}\label{ssec:survey-poor-standardisation}
Design choices in, e.g., stimulus creation, question formulation, stimulus presentation, and response format, can significantly affect the validity of human evaluation, and may introduce systematic biases. We find that current practices lack standardisation across these dimensions, and therefore it is generally infeasible to draw conclusions from cross-study comparisons.

\iffalse
\begin{figure}[t]
    \centering
    \includegraphics[width=\linewidth]{figure (4).pdf}
    \caption{\TODO{Update and clean.} Distribution of incompatible motion-quality attributes (left) and grounding aspects (right) utilised in contemporary evaluations.}
    \label{fig:review-question-formulations}
\end{figure}
\fi 
\paragraph{Question Formulations}
We observe an ad hoc variability in how evaluations formulate the concept of motion realism, with the usage of common adjectives shown on \cref{subfig:motion_quality_tags}. These terms lack a clear definition, and may encode different preferences -- for example, ``smooth'' or ``diverse'' motion does not necessarily assume that the motion is ``human-like'' or ``believable''. While different adjectives could theoretically be used to evaluate distinct styles (e.g., ``realism'' may be more desirable for photorealistic avatars than for cartoon characters), our survey finds this is not the case in practice. % Consequently, we believe that the inconsistency in question formulation needs to be addressed by the community, in order to avoid divergent and incomparable evaluation outcomes. 
% In practice, we believe that our evaluation of motion realism will correlate strongly with aspects such as visual quality, naturalness, ``fidelity'', and human-likeness. See also the results in \citet{kirkland2023stuck} that show that quality and naturalness, although not the same thing, are closely related for the case of text-to-speech systems trained on the same dataset.
The lack of standardisation extends to evaluations of multimodal alignment (\cref{subfig:grounding_factors}). Most evaluations attempt to isolate specific forms of alignment: expression of meaning and of style, or temporal alignment between speech and gesture, but we also find six publications evaluating general (unspecified) alignment to the speech. While the diversity of evaluation questions can reflect distinct modelling goals, the employed \emph{naive} evaluation setups may fail to isolate such nuances (cf.\ \cref{ssec:review-finding-disentanglement}).

%We therefore postulate that these nominally specialised evaluations in practice measure the same underlying construct of general alignment. This motivates our proposal to establish a standardised, general-purpose alignment task as a foundational measure quantified through mismatching (\cref{sec:evaluation guidelines}).

\begin{figure}
    \centering
    \includegraphics[width=\linewidth]{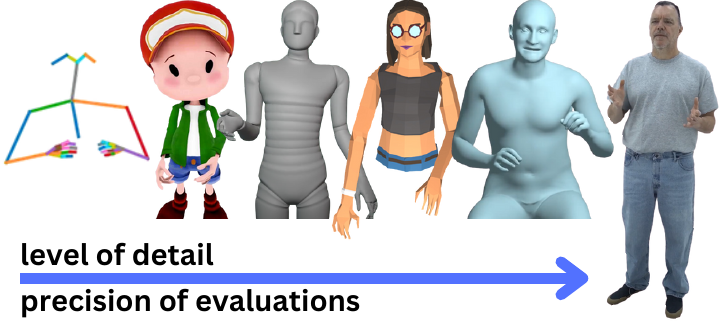}
    \caption{Example embodiments used in recent evaluations \cite{zhu2023taming, yang2023qpgesture, yoon2022genea, pang2023bodyformer, chhatre2024amuse, ng2024audio2photoreal}, highlighting the high variety in 3D characters and their degree of realism. There is empirical support for the idea that more expressive visualisations make performance differences substantially easier to spot \cite{ng2024audio2photoreal,du2025synthetically}.}
    %Evaluation results across gesture-generation publications are often not directly comparable due to the diversity of character visualisations used in user studies. (DiffGesture, SemanticGesticulator, SMPLX, RACER, QPGesture, Audio2Photoreal)}
    \label{fig:example-avatars-in-user-studies}
\end{figure}

\paragraph{Character Visualisation} Character visualisation is another critical factor in human evaluation design \cite{mcdonnell2012render}. Ideally, studies should use high-fidelity 3D avatars that can faithfully reproduce the original motion, avoiding artifacts from retargeting. Reinforcing this, a recent study by \citet{ng2024audio2photoreal} found that realistic rendering, compared to untextured meshes, leads to substantially more distinguishing power in evaluations.
%, as does presenting motion in VR \cite{du2025synthetically}.
Despite this, our survey finds that common practice falls short of this ideal. Several recent publications rely on simplistic stick-figure visualisations (\cite{qi2024emotransition, zhi2023livelyspeaker, cheng2025hop, zhu2023taming}), or untextured meshes (\cite{chhatre2024amuse, liu2024emage, yi2023generating, alexanderson2023listen, chen2024diffsheg}). The majority of surveyed evaluations % in \cref{tab:recent_models}
employ unique, non-standardised 3D characters with substantial differences in realism, artistic style, and even perceived personality traits (\cref{fig:example-avatars-in-user-studies}). These differences introduce yet another confounding variable to evaluations, undermining the comparability and generalisation of study results. 

\paragraph{User-Study Response Format} 
The literature is further fragmented by the diversity of methodologies for presenting stimuli and gathering responses. Taking the widely-used BEAT and BEAT2 datasets as an example, we find evaluations collecting  direct numerical ratings~\cite{qi2024emotransition, yang2023qpgesture}, pairwise forced choice votes between models~\cite{liu2024emage, mughal2024raggesture}, pairwise votes with ties~\cite{zhang2024semantic, ao2023gesturediffuclip, chhatre2024amuse}, and ordinal ranking of more than two videos~\cite{chen2024diffsheg, zhi2023livelyspeaker}. The fundamental problem is not that these methods would produce different model rankings, but that their results are not mutually comparable. For instance, the mean rating of a model on a 5-point scale cannot be compared to the win-rate of another in a pairwise setup. Ultimately, the lack of a common voting protocol prevents meaningful cross-study comparisons, obscuring the community's understanding of the true state of the art.

\subsection{Takeaways} \label{sec:review-takeaways}
Overall, we identified three critical shortcomings in current human evaluation practices in gesture generation:
\begin{enumerate}
    \item Doubtful ecological validity of evaluations of multimodal alignment, due to the uncontrolled confounding factor of motion realism.
    \item A systematic lack of direct comparisons between competing models.
    \item No standardisation of methodological design and other implementation factors of evaluation setups.
\end{enumerate}

Together, these three roadblocks make it impossible to reliably assess what the state of the art is, or understand which model is better for what purpose, without conducting the missing evaluations oneself. 
In fact, we theorise that current evaluations may be more performative than informative (cf.\ \citet{hertzmann2023curse}), possibly being more useful for paper acceptance than for actually measuring what they intend to measure.

\section{Evaluation Protocol for the BEAT2 Dataset} \label{sec:protocol}
\label{sec:evaluation guidelines}
Motivated by the findings of \cref{sec:limitations-survey}, we develop a human evaluation protocol for automatic 3D hand- and body gesture-generation models on the widely-used BEAT2 dataset \cite{liu2024emage}. Our goals are to establish standardised evaluations, and to provide a template that can be easily adapted and extended for new datasets and modelling problems. 

Using the GENEA Challenges as a starting point, our proposed protocol offers several methodological improvements to human evaluation of gesture generation. Later, in \cref{sec:experiments}, we use the proposed protocol to conduct a community-driven benchmark of six recent models.

\subsection{Disentangled Evaluation Dimensions}
Following general practice (cf. \cref{sec:limitations-survey}), we choose motion realism and speech-gesture alignment as the two evaluation dimensions of our protocol. Our review found that current human evaluation practices may lead to highly correlated measurements between the two dimensions (cf. \cref{ssec:review-finding-disentanglement}). This means that prior evaluations of methods claiming to improve, e.g., semantic gesturing, or temporal alignment, are unreliable due to the significant confounding effect of motion realism. To combat this problem, we propose a disentangled methodology similar to the GENEA Challenges \cite{kucherenko2021large, kucherenko2023genea, kucherenko2024evaluating}.

First, during motion-realism evaluations, we mute the audio in order to assess the visual quality of synthetic gestures in complete isolation from how well they're aligned with the speech. Second, to measure speech-gesture alignment, we employ a mismatching methodology where every model is only compared against its own (mismatched) outputs, completely removing the confounding effect of motion realism. Overall, these important design choices ensure that the two evaluation dimensions are fully independent. 

We remark that motion realism and speech-gesture alignment are not truly separable in human perception of gesturing. Whether motion looks natural depends on the context, and motion quality must be at a high enough level for motion to be appropriate. Nevertheless, it is crucial to separate the two aspects when evaluating synthetic gestures, because they are often targeted by distinct modelling choices.

\subsection{Motion-Realism Methodology} \label{ssec:protocol-motion-realism}
To evaluate motion realism, we pose the following question to user-study participants: \emph{``In which video does the character gesture more like a real person?''}.

In order to reduce the cognitive load of crowdsourced evaluators, we propose only using pairwise comparisons %. While potentially less time efficient than direct ratings (since each user response involves watching two video stimuli instead of just one), pairwise tests 
as they lead to higher inter-rater reliability \cite{wolfert2021rate}, which is crucial given the inherent difficulty of evaluating co-speech gesturing. For each pairwise comparison, we include five response options on a Likert-type scale (weak- or strong preference in either direction, and a tie). However, to capture detailed user feedback in an economical manner, we adapt the \emph{JUICE} methodology \cite{girdhar2024factorizing} -- originally for video generation -- to 3D motion evaluation for the first time. JUICE (``JUstify their choICE'') requires test-takers to not only indicate their preferences, but also select pre-defined reasons for their choices.
\subsubsection{Rating System}
The wide range of ranking mechanisms in gesture generation evaluations prevent cross-study comparisons (\cref{ssec:survey-poor-standardisation}). To solve this problem, we propose to use \textbf{Elo ratings} \cite{bradley1952rank} under the Bradley-Terry model \cite{bradley1952rank} as the standard ranking mechanism for motion realism in gesture generation. This methodology, originally for chess player ranking \cite{elo1967proposed}, has been successfully introduced to machine-learning evaluations by Chatbot Arena \cite{zheng2023chatbot_arena} (now LMArena).

A Bradley-Terry rating system assumes a fixed pairwise win-rate (under forced-choice comparison) for each pairing of gesture-generation models. Elo ratings are thereby computed using maximum-likelihood estimation such that the win-rate between two models $A$ and $B$ is a logistic sigmoid function of the difference between their respective Elo ratings $R_A$ and $R_B$ \cite{bradley1952rank}. Following common practice, we use a base-10 logistic function with a scaling factor $S=400$  such that:
\begin{equation}
P(\text{Model } A \text{ beats } \text{Model } B) = \frac{1}{1 + 10^{(R_B - R_A)/S}},
\end{equation}

Elo ratings offer several advantages. They are humanly interpretable: a difference of zero points means that the two systems are expected to beat each other half of the time (ignoring ties), whereas, e.g., a difference of +200 points (with our scaling factor) means that $A$ is expected to be judged as superior to $B$ in about 76\% of pairwise comparisons (averaged across all speech segments). Perhaps even more importantly, Elo ratings are inherently scalable as the number of evaluated models grows in a benchmark, since ranking between two models can be estimated from their Elo ratings even when direct comparisons are sparse or missing. 

%where $R_i$ is the current rating of model $i$, $R_j$ is the rating of the opponent model $j$, $E_i$ is the predicted probability that model $i$ wins against $j$, $S_i$ is the observed score (1 for a win, 0 for a loss, 0.5 for a tie), $R_i'$ is the updated rating of model $i$ after the match, $K$ is the learning factor controlling the update magnitude, and $\mathrm{BASE}$ and $\mathrm{SCALE}$ determine the logistic mapping from rating difference to win probability.

\subsection{Speech-Gesture Alignment Methodology}
\label{ssec:protocol-speech-gesture-alignment}
To evaluate speech-gesture alignment, we develop a mismatching protocol with the user-study question: \emph{``In which video do the character’s movements fit the speech better?''}. 

Similar to the motion-realism study, we combine five response options with additional JUICE feedback options detailed in the Appendix. Previous mismatching studies \cite{kucherenko2024evaluating, kucherenko2023genea} presented video clips with identical speech but different motion -- one matched, one mismatched with the audio. However, \citet{kucherenko2024evaluating} found that differences in motion-realism ratings between individual motion segments (regardless of whether they were matched to the speech) explained a significant fraction of user preferences also in mismatching studies. To address this, we propose \textbf{audio mismatching}, where each video in a pair has the exact same motion, but different speech audio: one matched, and one taken from another evaluation segment. This distinguishes our evaluation from prior works \cite{jonell2020let,rebol2021passing,yoon2022genea,kucherenko2023genea,kucherenko2024evaluating,yoon2020speech}.

By keeping motion constant between compared video clips, the realism of the motion can no longer bias the evaluators. Unlike all prior gesture-generation evaluations, this setup therefore fully disentangles the evaluation of motion realism (where only the visuals differ between stimulus pairs) from the evaluation of speech-gesture appropriateness (where the visuals are identical between the two stimuli in a pair). We note that there remains a potential bias between different speech segments, e.g., due to different voices being perceived as more appealing than others. We eliminate this bias by always using the same speaker voice in both stimuli. We further remove any systematic effect of possible preferences for individual speech segments within a speaker by ensuring that every speech segment appears equally often in matched as in mismatched videos.

\subsection{Additional details}
Please refer to \cref*{supp:protocol} for more information on our protocol, including test segment selection, user-study screens, crowdsourcing design, 3D visualisation, and the proposed JUICE factors for each evaluation.

\section{Experiments}
\label{sec:experiments}
To address the gap in the community's understanding of the state of the art in gesture generation (\cref{sec:limitations-survey}), and to validate our human evaluation protocol (\cref{sec:protocol}), we conduct a community-driven benchmarking effort. Due to known challenges with reproducibility in gesture generation, we invited authors of published models to participate by following our protocol, retraining their system under compatible settings, and submitting generated outputs for the BEAT2 English test set. To prevent cherry picking, we collect five sets of samples from each model (corresponding to five random seeds) for each clip in the test set.

In total, we compared six recently published gesture-
generation models --  DiffuseStyleGesture~\cite{yang2023diffusestylegesture}; Semantic Gesticulator~\cite{zhang2024semantic}; ConvoFusion~\cite{mughal2024convofusion}; RAG-Gesture~\cite{mughal2024raggesture}; AMUSE~\cite{chhatre2024amuse}; and HoloGest~\cite{cheng2025hologest} -- to each other and to the BEAT2 mocap recordings, recruiting over 600 evaluators and collecting over 16,000 pairwise votes on the \href{https://www.prolific.co/}{Prolific} crowdsourcing platform. For detailed information regarding the evaluated systems, test-taker demographics, user-study construction, JUICE analysis, and automatic metrics, we refer the reader to the Appendix.

%It is important to clarify that, although we filtered the initial speech segments to exclude instances where the corresponding motion-capture segment contains major artefacts, the motion of this condition does not always appear perfectly natural when visualised on a 3D avatar. This is primarily due to the fact that slight but consistent distortions of finger poses are present in large parts of the test set data, including in the motion corresponding to the speech segments selected for the leaderboard user studies.
\subsection{Motion-Realism Benchmark}
\label{ssec:realism-results}
We report the benchmark results obtained by following the motion realism protocol outlined in \cref{ssec:protocol-motion-realism}, and contrast them to prior results drawn from the status quo evaluation practices.

\begin{figure*}
    \centering
    \includegraphics[width=\linewidth]{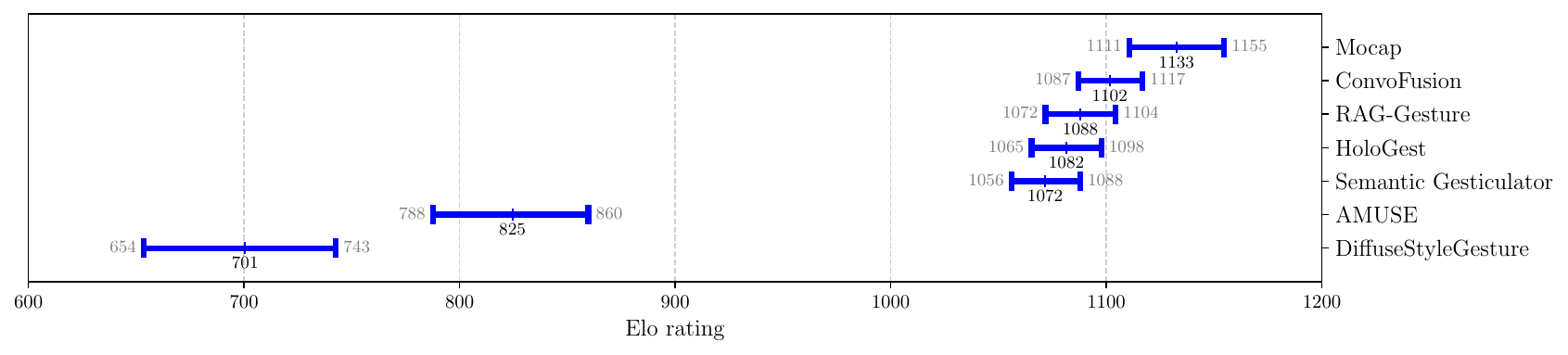}
    \caption{Results of the motion-realism user study, in the form of Elo ratings for each condition considered and 95\% confidence intervals acquired from bootstrapping. Conditions are ordered by estimated Elo rating in descending order.}
    \label{fig:realism-results}
\end{figure*}

\subsection{Prior Results on Motion Realism}
\textbf{ConvoFusion} and \textbf{RAG-Gesture} were evaluated with win rates in pairwise comparison, being preferred 43--47\% percent of the time against human motion \cite{mughal2024convofusion, mughal2024raggesture}. %%These claims are broadly consistent with their 30--45 point Elo-rating difference from the BEAT2 motion.
\textbf{HoloGest} used a five-point Likert-type rating scale in an evaluation with 30 test takers and reported average human-likeness scores of 4.61 for natural motion, 4.47 for HoloGest, and 3.70 for DiffuseStyleGesture \cite{cheng2025hologest}.
% HoloGest is in fact the only paper to have compared to another of the initial systems on the leaderboard, and their ordering matches our motion-realism evaluation.
The user study in \textbf{Semantic Gesticulator} \cite{zhang2024semantic} did not find their system to be significantly different from human motion. \textbf{AMUSE} \cite{chhatre2024amuse} outputs were tied with or preferred to motion capture 49\% of the time. \textbf{DiffuseStyleGesture} received an average score of about 4.1 on a five-point scale, statistically indistinguishable from natural motion at 4.2 in their user study \cite{yang2023diffusestylegesture}.

\begin{hlbox}
    In summary, every system was originally reported to be close to the visual quality of human motion capture, as measured by user studies. 
\end{hlbox}

\subsection{Our Results on Motion Realism}
On \cref{fig:realism-results}, we report Elo ratings with bootstrapped 95\% confidence intervals acquired from our motion realism protocol (\cref{ssec:protocol-motion-realism}). 
The \textbf{Mocap} condition is at the top of the results with a mean Elo rating of 1133, setting an empirical upper bound for motion realism in this study. Amongst the machine-learning systems, \textbf{ConvoFusion} and \textbf{RAG-Gesture} rank as the strongest models with Elo ratings of 1102 and 1088, respectively.
It makes sense for these systems to exhibit similar performance, given that RAG-Gesture builds on ConvoFusion. \textbf{HoloGest} follows closely behind RAG-Gesture with a mean Elo rating of 1084, slightly ahead of \textbf{Semantic Gesticulator} with an Elo rating of 1070.
All four of these systems remain well within a performance band that is arguably consistent with the broader state of the art.

At the lower end of the scale, \textbf{AMUSE} and \textbf{DiffuseStyleGesture} display notably lower Elo rating at 824 and 701, respectively. While AMUSE and DiffuseStyleGesture exhibit much wider confidence intervals, this does not imply that the comparative performance of these systems is less predictable. Rather, this is a consequence of early stopping in our adaptive evaluation due to their clear separation from other conditions.

Our results found a considerable gap between the best-performing gesture-generation models and AMUSE and DiffuseStyleGesture, even though both of those systems originally reported state-of-the-art performance. This may be taken as evidence that previous claims of high motion quality may not hold up under careful evaluation, or that faithfully adapting published models to a new dataset or to new dataset splits can be a significant challenge. Comparing the top systems, we find that:
\begin{hlbox}
    The BEAT2 dataset has become saturated for motion-realism evaluations, with four models showcasing comparable performance, with projected win rates between 41--46\% against motion-capture recordings.
\end{hlbox}

Importantly, this does not mean that the generated motion outputs are close to perfect. Rather, by intentionally removing the confounding factor of speech in our evaluations, we could show that the gap between gesture-generation models and natural human gesturing may be largely attributed to difficulties in aligning movements to the speech.

\subsection{\statusWIP{Speech-Gesture Alignment Benchmark}}
We also conduct the speech-gesture alignment evaluation as outlined in \cref{ssec:protocol-speech-gesture-alignment}. We report bootstrapped 95\% confidence intervals for audio mismatching scores, which are the weighted accuracies of preferring the matched stimulus, independently computed for each condition, with strong preference votes counting twice. For more details, please refer to the Appendix.
%\cref{sssec:appropriateness-statistics}.
\begin{figure}
    \centering
    \includegraphics[width=\linewidth]{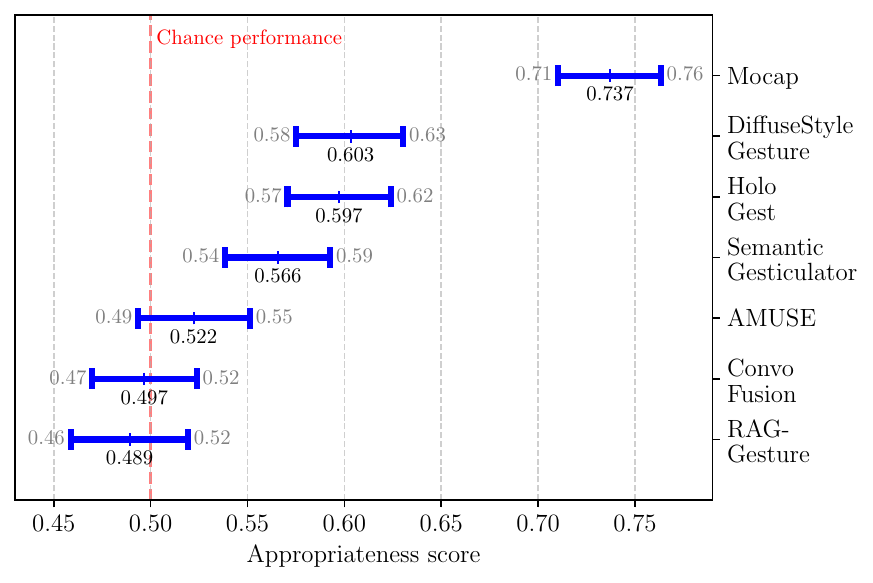}
    \caption{Results of the speech-gesture alignment user study. Ratio of user preference towards matched over mismatched stimuli for each condition considered, and 95\% confidence intervals acquired from bootstrapping. Conditions are ordered by estimated appropriateness score in descending order.}
    \label{fig:mismatching-results}
\end{figure}

\subsubsection{Prior Results on Speech-Gesture Alignment}
\textbf{DiffuseStyleGesture} performed a direct rating study which places its model at about an average score of 4.1 on a five-point scale, versus 4.2 for natural motion, calling it ``competitive'' with the latter \cite{yang2023diffusestylegesture}. The \textbf{HoloGest} paper again reports near-human mean opinion scores: 4.61 for human motion, 4.47 for their system, and DiffuseStyleGesture significantly behind at 3.91 \cite{cheng2025hologest}. \textbf{Semantic Gesticulator} reports a semantic accuracy score for BEAT data that is actually higher for their model (with a score of 0.41) than for human motion (with a score of 0.37), although the difference was not statistically significant \cite{zhang2024semantic}.
\textbf{ConvoFusion} found a mean semantic-alignment score of 2.97 on a five-point scale, compared to 3.53 for natural motion \cite{mughal2024convofusion}, whilst \textbf{RAG-Gesture} states that its output was preferred over natural motion 44.6 percent of the time and proposes that it sets the state of the art \cite{mughal2024raggesture}. Finally, \textbf{AMUSE} found a gesture-synchronisation win/loss rate of 35\%/52\% compared to test-set motion, with 12\% ties.

\begin{hlbox}
    Five out of six systems reported multimodal alignment results nearing or exceeding that of human recordings.
\end{hlbox}

\subsubsection{Our Results on Speech-Gesture Alignment} \label{ssec:our-alignment-results}
As shown in \cref{fig:mismatching-results}, our speech-gesture alignment study exposes significant differences compared to prior published evaluations. As expected, Mocap set the empirical topline with ${\approx}74\%$ mean audio mismatching score. DiffuseStyleGesture and HoloGest performed the best amongst generative models, both scoring ${\approx}60\%$. The near-identical alignment results suggest that HoloGest's previously reported alignment advances over DiffuseStyleGesture may in fact be confounded by motion quality, as postulated in \cref{ssec:review-finding-disentanglement}. Semantic Gesticulator reached ${\approx}57\%$, sharply contrasting prior results indicating strong semantic alignment. AMUSE, ConvoFusion, and RAG-Gesture all scored near 50\% mismatching scores, meaning that the motion they generate is only as valid for their input speech as for randomly chosen speech segments.

%all have confidence intervals overlapping $50\%$, indicating that their speech-gesture alignment of is not statistically significantly different from chance rate without FDR correction. This can be taken as evidence that their output could be interchanged for random gesture sequences in this task (or at least for random gesture sequences aligned to the start of a sentence, since that is how our segments were selected).

Comparing these results to the motion-realism study highlights surprising contrasts. First, prior claims (apart from possibly AMUSE) greatly differ from our findings, with ConvoFusion and RAG-Gesture performing at chance level. Second, system rankings shifted substantially. DiffuseStyleGesture, the weakest system in realism, now matches the top generative performer, whilst ConvoFusion and RAG-Gesture exhibited no measured effect of speech-gesture alignment. The reversal suggests that design choices that optimise kinematic plausibility do not necessarily translate to good grounding in the speech, and that a model that pays explicit attention to speech-gesture alignment can outperform more advanced motion models on this performance measure.

The nearly identical scores of ConvoFusion and RAG-Gesture are unexpected, given that RAG-Gesture was designed to enhance semantic appropriateness via retrieval-augmented generation. However, this mechanism does not yet offer a measurable benefit in our current evaluation. This may be due to factors such as the retrieved samples not aligning sufficiently with the input speech or the synthesis stage introducing noise that masks any semantic gain.
Another possibility is that the speech segments in our study may not provide enough opportunity for systems to demonstrate semantically grounded gesturing, as they were not pre-selected for this purpose.

Overall, we attribute the above differences between prior results and our findings to the standard evaluation practices failing to control for the confounding effect of motion realism, as discussed in \cref{ssec:review-finding-disentanglement}, artificially inflating the measured appropriateness, given the high visual quality many of the evaluated systems have. 

\begin{hlbox}
    The use of mismatching, as featured here, is critical in disentangling the quality of motion from its alignment with the speech. Prior evaluations of speech-gesture alignment are not reliable, and convey a false sense of progress, where in reality we have very far left to go.
\end{hlbox}

\section{Limitations}% and Future Work}
Our conclusions for state-of-the-art performance -- i.e., that motion realism is nearing saturation, while speech-gesture alignment is far from a solved problem -- are naturally limited by the scale and quality of BEAT2, and may not extend to other datasets. Adapting our evaluation protocol to other datasets is an important future direction.

\iffalse
The use of Elo ratings \cite{elo1967proposed} when evaluating motion realism comes with an assumption that the collective preferences of the test takers are transitive,
%(if A beats B and B beats C, then A must beat C),
which need not be the case; cf.\ \citet{arrow1950difficulty}.
\citet{boubdir2024elo} provides a discussion of this assumption in Elo-ratings-based evaluation of machine-learning models, along with possible alternatives.
That said, the vast majority of existing gesture evaluations surveyed in \cref{sec:limitations-survey} also assume transitivity, e.g., by having test takers assign ordinal ratings to different systems.
\fi 
Our mismatching-based evaluation of speech-gesture alignment is a diagnostic measure of specificity. If a gesture-generation system were to generate hypothetical ``universally appropriate outputs'' $x_{\mathrm{universal}}$ that are a good fit to any possible speech audio segment $s$, such a system will score close to chance level in a mismatching-based appropriateness study.
In practice, human motion contains rhythms at multiple levels \cite{pouw2021multilevel}, so a given motion $x$ can be perceived as rhythmically appropriate to a range of different speech audio with different rhythms (cf.\ \cite{miller2013you,dancingspiderman}), whereas semantic gestures (if produced by a system) are likely to be much more specific to a given context and speech.
This may, among other things, cause our evaluations to be relatively more sensitive to semantic gestures and their appropriateness over rhythmic (beat) appropriateness.
On the other hand, beats are vastly more common than semantic gestures, and are overall still likely to influence appropriateness evaluations more; cf.\ \citet{saund2021importance}.

Our literature review only considers speech-driven 3D hand- and body gesture generation. There are several closely related tasks -- e.g., solely text-driven generation, facial motion synthesis, or video generation -- that may suffer from similar problems, or offer alternative solutions to our methodology. 

We also note that some of the systems in our evaluation required adaptations, and re-training, for the BEAT2 dataset. Although we mitigated these effects by having original authors prepare the submissions, such changes may nevertheless have an adverse effect on model performance, We hope that the authors of future models will perform first-party evaluation using our protocol and our released data, for more accurate measurement of the state of the art.

Finally, the lack of reliable automatic metrics is a significant problem. The human preference responses collected during our evaluation are suitable for training human opinion predictors \cite{he2022automatic}, and may also enable the validation of existing and new automatic metrics \cite{Crnek2025Advancing,ismail2025establishing}.

\section{Conclusion}
In this paper, we surveyed systematic problems in gesture-generation evaluations and found that the state of the art cannot be determined from published results, and that the lack of standardisation may contribute to a false sense of progress. We then designed a new evaluation protocol, innovating on prior best practices with Elo ratings, audio mismatching, and JUICE. Finally, we put our protocol to the test by benchmarking six recent gesture-generation models, all of which have reported state-of-the-art performance.  

Our results provide a first-of-its-kind benchmark between competing published models in a field severely lacking direct comparisons. We find that motion realism on BEAT2 is no longer a distinctive factor between leading models, and -- perhaps even more importantly -- we show that current evaluation practices inflate multimodal-alignment results due to entanglement between evaluation dimensions. In contrast, our disentangled evaluations avoid common pitfalls, and pave the way towards reliable evaluations of more specialised aspects, e.g., semantic alignment or emotional expression.

We release all data collected in our evaluations (16,000 human votes, over 750 rendered video segments, and five hours of synthetic motion from the benchmarked models) to support standardised gesture-generation evaluations and the development of more targeted human evaluation methodologies and better automatic metrics.

\section*{Acknowledgements}
The authors thank Tu Anh Nguyen at FAIR for spotting a bug in an earlier version of our FGD evaluation code. RN, MT, TN, and GEH were partially supported by the Wallenberg AI, Autonomous Systems and Software Program (WASP) funded by the Knut and Alice Wallenberg Foundation. YY and GEH were partially supported by the Industrial Strategic Technology Development Program (grant no.\ 20023495) funded by MOTIE, Republic of Korea.

{
    \small
    \bibliographystyle{ieeenat_fullname}
    \bibliography{main}

@String{ACMToG = "ACM Trans. Graph." }

@String{ProcCVPR = "Proceedings of the IEEE/CVF Conference on Computer Vision and Pattern Recognition" }

@String{ProcICML = "Proceedings of the International Conference on Machine Learning" }

@String{ProcNeurIPS = "Advances in Neural Information Processing Systems" }

@String{ProcICCV = "Proceedings of the IEEE/CVF International Conference on Computer Vision" }

@String{ProcACMIVA = "Proceedings of the ACM International Conference on Intelligent Virtual Agents" }

@String{ProcECCV = "Proceedings of the European Conference on Computer Vision" }

@String{ProcICMI = "Proceedings of the ACM International Conference on Multimodal Interaction" }

@String{PRoySocB="P. Roy. Soc. B"}

@inproceedings{kucherenko2020gesticulator,
  title={Gesticulator: A framework for semantically-aware speech-driven gesture generation},
  author={Kucherenko, Taras and Jonell, Patrik and Van Waveren, Sanne and Henter, Gustav Eje and Alexandersson, Simon and Leite, Iolanda and Kjellstr{\"o}m, Hedvig},
  booktitle=ProcICMI,
  pages={242--250},
  year={2020}
}

@article{ao2023gesturediffuclip,
  title={Gesture{D}iffu{CLIP}: Gesture diffusion model with {CLIP} latents},
  author={Ao, Tenglong and Zhang, Zeyi and Liu, Libin},
  journal={ACM Transactions on Graphics (TOG)},
  volume={42},
  number={4},
  pages={1--18},
  year={2023},
  publisher={ACM New York, NY, USA}
}

@article{yoon2020speech,
  title={Speech gesture generation from the trimodal context of text, audio, and speaker identity},
  author={Yoon, Youngwoo and Cha, Bok and Lee, Joo-Haeng and Jang, Minsu and Lee, Jaeyeon and Kim, Jaehong and Lee, Geehyuk},
  journal={ACM Transactions on Graphics (TOG)},
  volume={39},
  number={6},
  pages={1--16},
  year={2020},
  publisher={ACM New York, NY, USA}
}

@inproceedings{alexanderson2020style,
  title={Style-controllable speech-driven gesture synthesis using normalising flows},
  author={Alexanderson, Simon and Henter, Gustav Eje and Kucherenko, Taras and Beskow, Jonas},
  booktitle={Computer Graphics Forum},
  volume={39},
  number={2},
  pages={487--496},
  year={2020},
  organization={Wiley Online Library}
}

@article{bradley1952rank,
  title={Rank analysis of incomplete block designs: {I}. The method of paired comparisons},
  author={Bradley, Ralph Allan and Terry, Milton E.},
  journal={Biometrika},
  volume={39},
  number={3/4},
  pages={324--345},
  year={1952}
}

@article{alexanderson2023listen,
  title={Listen, denoise, action! audio-driven motion synthesis with diffusion models},
  author={Alexanderson, Simon and Nagy, Rajmund and Beskow, Jonas and Henter, Gustav Eje},
  journal={ACM Transactions on Graphics (TOG)},
  volume={42},
  number={4},
  pages={1--20},
  year={2023},
  publisher={ACM New York, NY, USA}
}

@inproceedings{ghorbani2023zeroeggs,
  title={ZeroEGGS: Zero-shot Example-based Gesture Generation from Speech},
  author={Ghorbani, Saeed and Ferstl, Ylva and Holden, Daniel and Troje, Nikolaus F and Carbonneau, Marc-Andr{\'e}},
  booktitle={Computer Graphics Forum},
  volume={42},
  number={1},
  pages={206--216},
  year={2023},
  organization={Wiley Online Library}
}

@inproceedings{yang2023diffusestylegesture,
  title={Diffuse{S}tyle{G}esture: Stylized audio-driven co-speech gesture generation with diffusion models},
  author={Yang, Sicheng and Wu, Zhiyong and Li, Minglei and Zhang, Zhensong and Hao, Lei and Bao, Weihong and Cheng, Ming and Xiao, Long},
  booktitle={Proceedings of the International Joint Conference on Artificial Intelligence},
  pages={5860--5868},
  year={2023}
}

@inproceedings{zhu2023taming,
  title={Taming diffusion models for audio-driven co-speech gesture generation},
  author={Zhu, Lingting and Liu, Xian and Liu, Xuanyu and Qian, Rui and Liu, Ziwei and Yu, Lequan},
  booktitle={Proceedings of the IEEE/CVF Conference on Computer Vision and Pattern Recognition},
  pages={10544--10553},
  year={2023}
}

@inproceedings{liu2022learning,
  title={Learning hierarchical cross-modal association for co-speech gesture generation},
  author={Liu, Xian and Wu, Qianyi and Zhou, Hang and Xu, Yinghao and Qian, Rui and Lin, Xinyi and Zhou, Xiaowei and Wu, Wayne and Dai, Bo and Zhou, Bolei},
  booktitle=ProcCVPR,
  series={CVPR '22},
  pages={10462--10472},
  year={2022}
}

@article{pang2023bodyformer,
  title={Bodyformer: Semantics-guided 3d body gesture synthesis with transformer},
  author={Pang, Kunkun and Qin, Dafei and Fan, Yingruo and Habekost, Julian and Shiratori, Takaaki and Yamagishi, Junichi and Komura, Taku},
  journal={ACM Transactions on Graphics (TOG)},
  volume={42},
  number={4},
  pages={1--12},
  year={2023},
  publisher={ACM New York, NY, USA}
}

@inproceedings{yang2023qpgesture,
  title={{QPG}esture: Quantization-based and phase-guided motion matching for natural speech-driven gesture generation},
  author={Yang, Sicheng and Wu, Zhiyong and Li, Minglei and Zhang, Zhensong and Hao, Lei and Bao, Weihong and Zhuang, Haolin},
  booktitle={Proceedings of the IEEE/CVF Conference on Computer Vision and Pattern Recognition},
  pages={2321--2330},
  year={2023}
}

@inproceedings{nyatsanga_comprehensive_2023,
  title={A Comprehensive Review of Data-Driven Co-Speech Gesture Generation},
  author={Nyatsanga, Simbarashe and Kucherenko, Taras and Ahuja, Chaitanya and Henter, Gustav Eje and Neff, Michael},
  booktitle={Computer Graphics Forum},
  volume={42},
  number={2},
  pages={569--596},
  year={2023},
  organization={Wiley Online Library}
}

@inproceedings{lee2019talking,
  title={Talking {W}ith {H}ands 16.2 {M}: A large-scale dataset of synchronized body-finger motion and audio for conversational motion analysis and synthesis},
  author={Lee, Gilwoo and Deng, Zhiwei and Ma, Shugao and Shiratori, Takaaki and Srinivasa, Siddhartha S. and Sheikh, Yaser},
  booktitle={Proceedings of the IEEE/CVF International Conference on Computer Vision},
  pages={763--772},
  year={2019}
}

@inproceedings{liu2022beat,
  title={{BEAT}: A large-scale semantic and emotional multi-modal dataset for conversational gestures synthesis},
  author={Liu, Haiyang and Zhu, Zihao and Iwamoto, Naoya and Peng, Yichen and Li, Zhengqing and Zhou, You and Bozkurt, Elif and Zheng, Bo},
  booktitle={Proceedings of the European Conference on Computer Vision},
  pages={612--630},
  year={2022}
}

@inproceedings{ahuja2020no,
  title={No Gestures Left Behind: Learning Relationships between Spoken Language and Freeform Gestures},
  author={Ahuja, Chaitanya and Lee, Dong Won and Ishii, Ryo and Morency, Louis-Philippe},
  booktitle={Proceedings of the 2020 Conference on Empirical Methods in Natural Language Processing: Findings},
  pages={1884--1895},
  year={2020}
}

@article{wolfert2021review,
  title={A review of evaluation practices of gesture generation in embodied conversational agents},
  author={Wolfert, Pieter and Robinson, Nicole and Belpaeme, Tony},
  journal={IEEE Transactions on Human-Machine Systems},
  year={2022},
  publisher={IEEE},
  volume={52},
  number={3},
  pages={379-389},
  doi={10.1109/THMS.2022.3149173}
}

@inproceedings{yoon2022genea,
  title={The GENEA Challenge 2022: A large evaluation of data-driven co-speech gesture generation},
  author={Yoon, Youngwoo and Wolfert, Pieter and Kucherenko, Taras and Viegas, Carla and Nikolov, Teodor and Tsakov, Mihail and Henter, Gustav Eje},
  booktitle={Proceedings of the 2022 International Conference on Multimodal Interaction},
  pages={736--747},
  year={2022}
}

@article{kucherenko2024evaluating,
  title={Evaluating gesture generation in a large-scale open challenge: The {GENEA} {C}hallenge 2022},
  author={Kucherenko, Taras and Wolfert, Pieter and Yoon, Youngwoo and Viegas, Carla and Nikolov, Teodor and Tsakov, Mihail and Henter, Gustav Eje},
  journal={ACM Transactions on Graphics (TOG)},
  doi={10.1145/3656374},
  year={2024}
}

@inproceedings{li2021ai,
  title={{AI} choreographer: Music conditioned 3{D} dance generation with {AIST}++},
  author={Li, Ruilong and Yang, Shan and Ross, David A. and Kanazawa, Angjoo},
  booktitle=ProcCVPR,
  series={CVPR '21},
  pages={13401--13412},
  year={2021}
}

@inproceedings{qi2024emotransition,
    author    = {Qi, Xingqun and Pan, Jiahao and Li, Peng and Yuan, Ruibin and Chi, Xiaowei and Li, Mengfei and Luo, Wenhan and Xue, Wei and Zhang, Shanghang and Liu, Qifeng and Guo, Yike},
    title     = {Weakly-Supervised Emotion Transition Learning for Diverse 3D Co-speech Gesture Generation},
    booktitle = ProcCVPR,
    series={CVPR '24},
    month     = {June},
    year      = {2024},
    pages     = {10424-10434}
}

@InProceedings{mughal2024convofusion,
title = {ConvoFusion: Multi-Modal Conversational Diffusion for Co-Speech Gesture Synthesis},
author = {Muhammad Hamza Mughal and Rishabh Dabral and Ikhsanul Habibie and Lucia Donatelli and Marc Habermann and Christian Theobalt},
booktitle=ProcCVPR,
series={CVPR '24},
year={2024}
}

@InProceedings{tseng2023edge,
    author    = {Tseng, Jonathan and Castellon, Rodrigo and Liu, Karen},
    title     = {{EDGE}: Editable Dance Generation From Music},
    booktitle=ProcCVPR,
    series={CVPR '23},
    month     = {June},
    year      = {2023},
    pages     = {448-458}
}

@inproceedings{ng2024audio2photoreal,
  title={From Audio to Photoreal Embodiment: Synthesizing Humans in Conversations},
  author={Ng, Evonne and Romero, Javier and Bagautdinov, Timur and Bai, Shaojie and Darrell, Trevor and Kanazawa, Angjoo and Richard, Alexander},
  booktitle={IEEE Conference on Computer Vision and Pattern Recognition},
  year={2024}
}

@InProceedings{ahuja2023continual,
    author    = {Ahuja, Chaitanya and Joshi, Pratik and Ishii, Ryo and Morency, Louis-Philippe},
    title     = {Continual Learning for Personalized Co-speech Gesture Generation},
    booktitle = {Proceedings of the IEEE/CVF International Conference on Computer Vision (ICCV)},
    month     = {October},
    year      = {2023},
    pages     = {20893-20903}
}

@InProceedings{chhatre2024amuse,
    author    = {Chhatre, Kiran and Daněček, Radek and Athanasiou, Nikos and Becherini, Giorgio and Peters, Christopher and Black, Michael J. and Bolkart, Timo},
    title     = {{AMUSE}: Emotional Speech-driven {3D} Body Animation via Disentangled Latent Diffusion},
    booktitle=ProcCVPR,
    series={CVPR '24},
    month     = {June},
    year      = {2024},
    pages     = {1942-1953},
    url = {https://amuse.is.tue.mpg.de},
}

@inproceedings{liu2024towards_probtalk,
  title={Towards variable and coordinated holistic co-speech motion generation},
  author={Liu, Yifei and Cao, Qiong and Wen, Yandong and Jiang, Huaiguang and Ding, Changxing},
  booktitle={Proceedings of the IEEE/CVF Conference on Computer Vision and Pattern Recognition},
  pages={1566--1576},
  year={2024}
}

@inproceedings{chen2024diffsheg,
  title     = {DiffSHEG: A Diffusion-Based Approach for Real-Time Speech-driven Holistic 3D Expression and Gesture Generation},
  author    = {Chen, Junming and Liu, Yunfei and Wang, Jianan and Zeng, Ailing and Li, Yu and Chen, Qifeng},
  booktitle=ProcCVPR,
  series={CVPR '24},
  year      = {2024}
}

@inproceedings{kucherenko2023genea,
  title={The {GENEA} {C}hallenge 2023: A large-scale evaluation of gesture generation models in monadic and dyadic settings},
  author={Kucherenko, Taras and Nagy, Rajmund and Yoon, Youngwoo and Woo, Jieyeon and Nikolov, Teodor and Tsakov, Mihail and Henter, Gustav Eje},
  booktitle={Proceedings of the International Conference on Multimodal Interaction},
  pages={792--801},
  year={2023}
}

@inproceedings{kucherenko2021large,
  title={A large, crowdsourced evaluation of gesture generation systems on common data: The GENEA Challenge 2020},
  author={Kucherenko, Taras and Jonell, Patrik and Yoon, Youngwoo and Wolfert, Pieter and Henter, Gustav Eje},
  booktitle={26th international conference on intelligent user interfaces},
  pages={11--21},
  year={2021}
}

@InProceedings{dabral2022mofusion,
      title={MoFusion: A Framework for Denoising-Diffusion-based Motion Synthesis},
      author={Rishabh Dabral and Muhammad Hamza Mughal and Vladislav Golyanik and Christian Theobalt},
      booktitle=ProcCVPR,
      series={CVPR '23},
      year={2023}
}

@inproceedings{zheng2023chatbot_arena,
  title={Judging {LLM}-as-a-judge with {MT}-{B}ench and {C}hatbot {A}rena},
  author={Zheng, Lianmin and Chiang, Wei-Lin and Sheng, Ying and Zhuang, Siyuan and Wu, Zhanghao and Zhuang, Yonghao and Lin, Zi and Li, Zhuohan and Li, Dacheng and Xing, Eric and others},
  booktitle=ProcNeurIPS,
  series={NeurIPS '23},
  pages={46595--46623},
  volume={36},
  year={2023}
}

@inproceedings{deichler2023diffusion,
  title={Diffusion-based co-speech gesture generation using joint text and audio representation},
  author={Deichler, Anna and Mehta, Shivam and Alexanderson, Simon and Beskow, Jonas},
  booktitle={Proceedings of the International Conference on Multimodal Interaction},
  pages={755--762},
  year={2023}
}

@inproceedings{perrotin2023blizzard,
  title={The Blizzard Challenge 2023},
  author={Perrotin, Olivier and Stephenson, Brooke and Gerber, Silvain and Bailly, G{\'e}rard},
  booktitle={18th Blizzard Challenge Workshop},
  pages={1--27},
  year={2023},
  organization={ISCA}
}

@inproceedings{yazdian2022gesture2vec,
  title={{G}esture2{V}ec: Clustering gestures using representation learning methods for co-speech gesture generation},
  author={Yazdian, Payam Jome and Chen, Mo and Lim, Angelica},
  booktitle={Proceedings of the IEEE/RSJ International Conference on Intelligent Robots and Systems},
  series={IROS '22},
  pages={3100--3107},
  year={2022}
}

@inproceedings{sun2023co,
  title={Co-speech Gesture Synthesis by Reinforcement Learning with Contrastive Pre-trained Rewards},
  author={Sun, Mingyang and Zhao, Mengchen and Hou, Yaqing and Li, Minglei and Xu, Huang and Xu, Songcen and Hao, Jianye},
  booktitle={Proceedings of the IEEE/CVF Conference on Computer Vision and Pattern Recognition},
  pages={2331--2340},
  year={2023}
}

@inproceedings{zhi2023livelyspeaker,
  title={LivelySpeaker: Towards semantic-aware co-speech gesture generation},
  author={Zhi, Yihao and Cun, Xiaodong and Chen, Xuelin and Shen, Xi and Guo, Wen and Huang, Shaoli and Gao, Shenghua},
  booktitle={Proceedings of the IEEE/CVF International Conference on Computer Vision},
  pages={20807--20817},
  year={2023}
}

@inproceedings{yi2023generating,
  title={Generating holistic 3d human motion from speech},
  author={Yi, Hongwei and Liang, Hualin and Liu, Yifei and Cao, Qiong and Wen, Yandong and Bolkart, Timo and Tao, Dacheng and Black, Michael J},
  booktitle={Proceedings of the IEEE/CVF Conference on Computer Vision and Pattern Recognition},
  pages={469--480},
  year={2023}
}

@inproceedings{yoon2019robots,
  title={Robots learn social skills: End-to-end learning of co-speech gesture generation for humanoid robots},
  author={Yoon, Youngwoo and Ko, Woo-Ri and Jang, Minsu and Lee, Jaeyeon and Kim, Jaehong and Lee, Geehyuk},
  booktitle={2019 International Conference on Robotics and Automation (ICRA)},
  pages={4303--4309},
  year={2019},
  organization={IEEE}
}

@inproceedings{saund2021importance,
  title={The Importance of Qualitative Elements in Subjective Evaluation of Semantic Gestures},
  author={Saund, Carolyn and Marsella, Stacy},
  booktitle={2021 16th IEEE International Conference on Automatic Face and Gesture Recognition (FG 2021)},
  pages={1--8},
  year={2021},
  organization={IEEE}
}

@article{ennis2010seeing,
  title={Seeing is believing: body motion dominates in multisensory conversations},
  author={Ennis, Cathy and McDonnell, Rachel and O'Sullivan, Carol},
  journal={ACM Transactions on Graphics (TOG)},
  volume={29},
  number={4},
  pages={1--9},
  year={2010}
}

@inproceedings{cooper2023investingating,
  author={Cooper, Erica and Yamagishi, Junichi},
  title={Investigating Range-Equalizing Bias in Mean Opinion Score Ratings of Synthesized Speech},
  year={2023},
  booktitle={Proc. Interspeech},
  pages={1104--1108}
}

@InProceedings{liu2024emage,
    author    = {Liu, Haiyang and Zhu, Zihao and Becherini, Giorgio and Peng, Yichen and Su, Mingyang and Zhou, You and Zhe, Xuefei and Iwamoto, Naoya and Zheng, Bo and Black, Michael J.},
    title     = {{EMAGE}: Towards Unified Holistic Co-Speech Gesture Generation via Expressive Masked Audio Gesture Modeling},
  booktitle=ProcCVPR,
  series={CVPR '24},
    month     = {June},
    year      = {2024},
    pages     = {1144-1154}
}

@inproceedings{wolfert2021rate,
  author = {Wolfert, Pieter and Girard, Jeffrey M. and Kucherenko, Taras and Belpaeme, Tony},
  title = {To rate or not to rate: Investigating evaluation methods for generated co-speech gestures},
  year = {2021},
  publisher = {ACM},
  doi = {10.1145/3462244.3479889},
  booktitle = {Proc. ICMI},
  pages = {494--502},
  numpages = {9},
  series = {ICMI '21}
}

@article{he2022automatic,
  title={Automatic quality assessment of speech-driven synthesized gestures},
  author={He, Zhiyuan},
  journal={International Journal of Computer Games Technology},
  volume={2022},
  year={2022},
  doi={10.1155/2022/1828293}
}

@article{hertzmann2023curse,
  title={The Curse of Performative User Studies},
  author={Hertzmann, Aaron},
  journal={IEEE Computer Graphics and Applications},
  volume={43},
  number={6},
  pages={112--116},
  year={2023},
  publisher={IEEE}
}

@inproceedings{rebol2021passing,
  title={Passing a non-verbal {T}uring test: Evaluating gesture animations generated from speech},
  author={Rebol, Manuel and G{\"u}ti, Christian and Pietroszek, Krzysztof},
  booktitle={Proceedings of the IEEE Conference on Virtual Reality and 3D User Interfaces},
  pages={573--581},
  year={2021},
  publisher={IEEE},
  doi={10.1109/VR50410.2021.00082},
  series={VR '21}
}

@inproceedings{chong2020effectively,
  title={Effectively unbiased {FID} and {I}nception score and where to find them},
  author={Chong, Min Jin and Forsyth, David},
  booktitle=ProcCVPR,
  series={CVPR '20},
  pages={6070--6079},
  year={2020}
}

@inproceedings{pavlakos2019cvpr,
  title = {Expressive Body Capture: {3D} Hands, Face, and Body from a Single Image},
  author = {Pavlakos, Georgios and Choutas, Vasileios and Ghorbani, Nima and Bolkart, Timo and Osman, Ahmed A. A. and Tzionas, Dimitrios and Black, Michael J.},
  booktitle=ProcCVPR,
  series={CVPR '19},
  pages     = {10975--10985},
  year = {2019}
}

@inproceedings{
chen2024language,
  title={The Language of Motion: Unifying Verbal and Non-verbal Language of 3D Human Motion},
  author={Changan Chen and Juze Zhang and Shrinidhi Kowshika Lakshmikanth and Yusu Fang and Ruizhi Shao and Gordon Wetzstein and Li Fei-Fei and Ehsan Adeli},
  booktitle=ProcCVPR,
  series={CVPR '25},
year={2025}
}

@InProceedings{mughal2024raggesture,
title = {Retrieving Semantics from the Deep: an RAG Solution for Gesture Synthesis},
author = {M. Hamza Mughal and Rishabh Dabral and Merel C. J. Scholman and Vera Demberg and Christian Theobalt},
booktitle=ProcCVPR,
series={CVPR '25},
year={2025}
}

@inproceedings{cheng2025hop,
  title     = {HOP: Heterogeneous Topology-based Multimodal Entanglement for Co-Speech Gesture Generation},
  author    = {Cheng, Hongye and Wang, Tianyu and Shi, Guangsi and Zhao, Zexing and Fu, Yanwei},
  booktitle=ProcCVPR,
  series={CVPR '25},
  year={2025}
}

@inproceedings{voss2023aq,
  title={AQ-GT: a temporally aligned and quantized GRU-transformer for co-speech gesture synthesis},
  author={Vo{\ss}, Hendric and Kopp, Stefan},
  booktitle={Proceedings of the 25th International Conference on Multimodal Interaction},
  pages={60--69},
  year={2023}
}

@article{bylinskii2023towards,
  title={Towards better user studies in computer graphics and vision},
  author={Bylinskii, Zoya and Herman, Laura and Hertzmann, Aaron and Hutka, Stefanie and Zhang, Yile and others},
  journal={Foundations and Trends{\textregistered} in Computer Graphics and Vision},
  volume={15},
  number={3},
  pages={201--252},
  year={2023},
  publisher={Now Publishers, Inc.}
}

@inproceedings{takeuchi2017speech,
  title={Speech-to-Gesture Generation: A Challenge in Deep Learning Approach with Bi-Directional {LSTM}},
  author={Takeuchi, Kenta and Hasegawa, Dai and Shirakawa, Shinichi and Kaneko, Naoshi and Sakuta, Hiroshi and Sumi, Kazuhiko},
  booktitle={Proceedings of the International Conference on Human Agent Interaction},
  year={2017},
  series={HAI '17}
}

@inproceedings{hasegawa2018evaluation,
  title={Evaluation of speech-to-gesture generation using bi-directional {LSTM} network},
  author={Hasegawa, Dai and Kaneko, Naoshi and Shirakawa, Shinichi and Sakuta, Hiroshi and Sumi, Kazuhiko},
  booktitle=ProcACMIVA,
  series = {IVA'18},
  publisher = {ACM},
  address = {New York, NY, USA},
  doi = {10.1145/3267851.3267878},
  numpages = {8},
  pages={79--86},
  year={2018}
}

@inproceedings{ferstl2018trinity,
  title={Investigating the use of recurrent motion modelling for speech gesture generation},
  author={Ferstl, Ylva and McDonnell, Rachel},
  booktitle = ProcACMIVA,
  pages = {93--98},
  year={2018},
  url = {https://trinityspeechgesture.scss.tcd.ie}
}

@inproceedings{kucherenko2019analyzing,
  author={Kucherenko, Taras and Hasegawa, Dai and Henter, Gustav Eje and Kaneko, Naoshi and Kjellstr{\"o}m, Hedvig},
  booktitle=ProcACMIVA,
  series = {IVA'19},
  publisher = {ACM},
  address = {New York, NY, USA},
  pages={97--104},
  numpages = {8},
  title={Analyzing input and output representations for speech-driven gesture generation},
  doi = {10.1145/3308532.3329472},
  year={2019}
}

@article{kucherenko2021moving,
  title={Moving fast and slow: {A}nalysis of representations and post-processing in speech-driven automatic gesture generation},
  author={Kucherenko, Taras and Hasegawa, Dai and Kaneko, Naoshi and Henter, Gustav Eje and Kjellstr{\"o}m, Hedvig},
  journal={International Journal of Human-Computer Interaction},
  publisher={Taylor \& Francis},
  volume={37},
  number={14},
  pages={1300--1316},
  doi={10.1080/10447318.2021.1883883},
  year={2021}
}

@inproceedings{rombach2022high,
  title={High-resolution image synthesis with latent diffusion models},
  author={Rombach, Robin and Blattmann, Andreas and Lorenz, Dominik and Esser, Patrick and Ommer, Bj{\"o}rn},
  booktitle=ProcCVPR,
  pages={10684--10695},
  year={2022},
  series={CVPR '22}
}

@inproceedings{jonell2020let,
  title={Let's face it: Probabilistic multi-modal interlocutor-aware generation of facial gestures in dyadic settings},
  author={Jonell, Patrik and Kucherenko, Taras and Henter, Gustav Eje and Beskow, Jonas},
  booktitle=ProcACMIVA,
  articleno = {31},
  numpages = {8},
  series={IVA '20},
  doi={10.1145/3383652.3423911},
  year={2020}
}

@article{elo1967proposed,
  title={The proposed {USCF} rating system, its development, theory, and applications},
  author={Elo, Arpad E.},
  journal={Chess Life},
  volume={22},
  number={8},
  pages={242--247},
  year={1967}
}

@article{boubdir2024elo,
  title={Elo uncovered: Robustness and best practices in language model evaluation},
  author={Boubdir, Meriem and Kim, Edward and Ermis, Beyza and Hooker, Sara and Fadaee, Marzieh},
  journal={Advances in Neural Information Processing Systems},
  volume={37},
  pages={106135--106161},
  year={2024}
}

@inproceedings{girdhar2024factorizing,
  title={Factorizing text-to-video generation by explicit image conditioning},
  author={Girdhar, Rohit and Singh, Mannat and Brown, Andrew and Duval, Quentin and Azadi, Samaneh and Rambhatla, Sai Saketh and Shah, Akbar and Yin, Xi and Parikh, Devi and Misra, Ishan},
  booktitle=ProcECCV,
  pages={205--224},
  year={2024}
}

@inproceedings{liu2021speech,
  title={Speech-based gesture generation for robots and embodied agents: A scoping review},
  author={Liu, Yu and Mohammadi, Gelareh and Song, Yang and Johal, Wafa},
  booktitle={Proceedings of the International Conference on Human-Agent Interaction},
  pages={31--38},
  year={2021},
  series = {HAI '21}
}

@article{abootorabi2025generative,
  title={Generative {AI} for Character Animation: A Comprehensive Survey of Techniques, Applications, and Future Directions},
  author={Abootorabi, Mohammad Mahdi and Ghahroodi, Omid and Zahraei, Pardis Sadat and Behzadasl, Hossein and Mirrokni, Alireza and Salimipanah, Mobina and Rasouli, Arash and Behzadipour, Bahar and Azarnoush, Sara and Maleki, Benyamin and others},
  journal={arXiv preprint arXiv:2504.19056},
  year={2025}
}

@book{kendall1948rank,
  title={Rank correlation methods.},
  author={Kendall, Maurice George},
  year={1948},
  publisher={Griffin}
}

@article{arrow1950difficulty,
  title={A difficulty in the concept of social welfare},
  author={Arrow, Kenneth J.},
  journal={Journal of Political Economy},
  volume={58},
  number={4},
  pages={328--346},
  year={1950}
}

@article{pouw2021multilevel,
  title={Multilevel rhythms in multimodal communication},
  author={Pouw, Wim and Proksch, Shannon and Drijvers, Linda and Gamba, Marco and Holler, Judith and Kello, Christopher and Schaefer, Rebecca S. and Wiggins, Geraint A.},
  journal=PRoySocB,
  volume={376},
  number={1835},
  articleno={20200334},
  year={2021}
}

@article{miller2013you,
  title={When what you hear influences when you see: listening to an auditory rhythm influences the temporal allocation of visual attention},
  author={Miller, Jared E. and Carlson, Laura A. and McAuley, J. Devin},
  journal={Psychological Science},
  volume={24},
  number={1},
  pages={11--18},
  year={2013}
}

@misc{dancingspiderman,
   author =       {Jody Avirgan},
   title =        {Why Spiderman is such a good dancer},
   editor =       {WNYC Studios},
   month =        {June},
   year =         {2013},
   howpublished = {\url{https://web.archive.org/web/20201112011116/https://www.wnycstudios.org/podcasts/radiolab/articles/299399-why-spiderman-such-good-dancer}},
 }

@article{chatziagapi2025avflow,
  title={{AV-F}low: Transforming text to audio-visual human-like interactions},
  author={Aggelina Chatziagapi and Louis-Philippe Morency and Hongyu Gong and Michael Zollh{\"o}fer and Dimitris Samaras and Alexander Richard},
  year={2025},
  journal={arXiv preprint arXiv:2502.13133}
}

@misc{chiang2023update,
  title = {{C}hatbot {A}rena: New models \& {E}lo system update},
  author = {Wei-Lin Chiang and Tim Li and Joseph E. Gonzalez and Ion Stoica},
  howpublished = {\url{https://lmsys.org/blog/2023-12-07-leaderboard/}},
  year={2023},
  note = {Accessed: 2025-05-20}
}

@article{zhang2024semantic,
author = {Zhang, Zeyi and Ao, Tenglong and Zhang, Yuyao and Gao, Qingzhe and Lin, Chuan and Chen, Baoquan and Liu, Libin},
title = {Semantic Gesticulator: Semantics-aware co-speech gesture synthesis},
year = {2024},
volume = {43},
number = {4},
doi = {10.1145/3658134},
journal = ACMToG,
articleno = {136},
numpages = {17}
}

@inproceedings{cheng2025hologest,
  title={{H}olo{G}est: Decoupled diffusion and motion priors for generating holisticly expressive co-speech gestures},
  author={Cheng, Yongkang and Huang, Shaoli},
  booktitle={Proceedings of the International Conference on 3D Vision},
  series={3DV '25},
  url={https://arxiv.org/abs/2503.13229},
  year={2025}
}

@InProceedings{pmlr-v139-touvron21a,
  title = 	 {Training data-efficient image transformers \& distillation through attention},
  author =       {Touvron, Hugo and Cord, Matthieu and Douze, Matthijs and Massa, Francisco and Sablayrolles, Alexandre and Jegou, Herve},
  booktitle = 	 ProcICML,
  pages = 	 {10347--10357},
  year = 	 {2021},
  editor = 	 {Meila, Marina and Zhang, Tong},
  volume = 	 {139},
  series = 	 {Proceedings of Machine Learning Research},
  month = 	 {18--24 Jul},
  publisher =    {PMLR},
  url = 	 {https://proceedings.mlr.press/v139/touvron21a.html}
}

@article{baevski2020wav2vec,
  title={wav2vec 2.0: A framework for self-supervised learning of speech representations},
  author={Baevski, Alexei and Zhou, Yuhao and Mohamed, Abdelrahman and Auli, Michael},
  journal={Advances in neural information processing systems},
  volume={33},
  pages={12449--12460},
  year={2020}
}

@inproceedings{haque2025wild,
  title={“Wild West” of Evaluating Speech-Driven 3D Facial Animation Synthesis: A Benchmark Study},
  author={Haque, Kazi Injamamul and Pavlou, Alkiviadis and Yumak, Zerrin},
  booktitle={Computer Graphics Forum},
  pages={e70073},
  year={2025},
  organization={Wiley Online Library}
}

@inproceedings{yang2024probabilistic,
  title={Probabilistic speech-driven 3D facial motion synthesis: new benchmarks methods and applications},
  author={Yang, Karren D and Ranjan, Anurag and Chang, Jen-Hao Rick and Vemulapalli, Raviteja and Tuzel, Oncel},
  booktitle={Proceedings of the IEEE/CVF Conference on Computer Vision and Pattern Recognition},
  pages={27294--27303},
  year={2024}
}

@article{mcdonnell2012render,
  title={Render me real? Investigating the effect of render style on the perception of animated virtual humans},
  author={McDonnell, Rachel and Breidt, Martin and B{\"u}lthoff, Heinrich H},
  journal={ACM Transactions on Graphics (TOG)},
  volume={31},
  number={4},
  pages={1--11},
  year={2012},
  publisher={ACM New York, NY, USA}
}

@inproceedings{yang2023diffusestylegesture+,
  title={The DiffuseStyleGesture+ entry to the GENEA Challenge 2023},
  author={Yang, Sicheng and Xue, Haiwei and Zhang, Zhensong and Li, Minglei and Wu, Zhiyong and Wu, Xiaofei and Xu, Songcen and Dai, Zonghong},
  booktitle={Proceedings of the 25th International Conference on Multimodal Interaction},
  pages={779--785},
  year={2023}
}

@article{Crnek2025Advancing,
author = {Karlo Crnek and Grega Močnik and Matej Rojc},
title = {Advancing Objective Evaluation of Speech-Driven Gesture Generation for Embodied Conversational Agents},
journal = {International Journal of Human–Computer Interaction},
volume = {0},
number = {0},
pages = {1--17},
year = {2025},
publisher = {Taylor \& Francis},
doi = {10.1080/10447318.2025.2531286}
}

@article{ismail2025establishing,
  title={Establishing a unified evaluation framework for human motion generation: A comparative analysis of metrics},
  author={Ismail-Fawaz, Ali and Devanne, Maxime and Berretti, Stefano and Weber, Jonathan and Forestier, Germain},
  journal={Computer Vision and Image Understanding},
  volume={254},
  pages={104337},
  year={2025},
  publisher={Elsevier}
}

@inproceedings{yang2025GestureHYDRA,
  author    = {Quanwei Yang and Luying Huang and Kaisiyuan Wang and Jiazhi Guan and Shengyi He and Fengguo Li and Lingyun Yu and Yingying Li and Haocheng Feng and Hang Zhou and Hongtao Xie.},
  title     = {GestureHYDRA: Semantic Co-speech Gesture Synthesis via Hybrid Modality Diffusion Transformer and Cascaded-Synchronized Retrieval-Augmented Generation},
    booktitle=ProcICCV,
    series={ICCV '25},
    year={2025}
}

@inproceedings{liu2025semges,
    title     = {SemGes: Semantics-aware Co-Speech Gesture Generation using Semantic Coherence and Relevance Learning},
    author    = {Lanmiao Liu and Esam Ghaleb and Aslı Özyürek and Zerrin Yumak},
    year      = {2025},
    booktitle=ProcICCV,
    series={ICCV '25},
}

@inproceedings{liu2025gesturelsmlatentshortcutbased,
  title={GestureLSM: Latent Shortcut based Co-Speech Gesture Generation with Spatial-Temporal Modeling},
  author={Pinxin Liu and Luchuan Song and Junhua Huang and Chenliang Xu},
    booktitle=ProcICCV,
    series={ICCV '25},
    year={2025}
}

@inproceedings{zhang2024semtalk,
  title={SemTalk: Holistic Co-speech Motion Generation with Frame-level Semantic Emphasis},
  author={Zhang, Xiangyue and Li, Jianfang and Zhang, Jiaxu and Dang, Ziqiang and Ren, Jianqiang and Bo, Liefeng and Tu, Zhigang},
    booktitle=ProcICCV,
    series={ICCV '25},
    year={2025}
}

@inproceedings{chen2025meco,
  author = {Bohong Chen and Yumeng Li and Youyi Zheng and Yao-Xiang Ding and Kun Zhou},
  title = {Motion-example-controlled Co-speech Gesture Generation Leveraging Large Language Models},
  year = {2025},
  isbn = {9798400715402},
  publisher = {Association for Computing Machinery},
  address = {New York, NY, USA},
  url = {https://doi.org/10.1145/3721238.3730611},
  doi = {10.1145/3721238.3730611},
  booktitle = {Proceedings of the Special Interest Group on Computer Graphics and Interactive Techniques Conference Conference Papers},
  series = {SIGGRAPH Conference Papers '25}
}

@inproceedings{cheng2024siggesture,
author = {Cheng, Qingrong and Li, Xu and Fu, Xinghui},
title = {SIGGesture: Generalized Co-Speech Gesture Synthesis via Semantic Injection with Large-Scale Pre-Training Diffusion Models},
year = {2024},
isbn = {9798400711312},
publisher = {Association for Computing Machinery},
address = {New York, NY, USA},
url = {https://doi.org/10.1145/3680528.3687677},
doi = {10.1145/3680528.3687677},
booktitle = {SIGGRAPH Asia 2024 Conference Papers},
articleno = {133},
numpages = {11},
location = {Tokyo, Japan},
series = {SA '24}
}

@inproceedings{du2025synthetically,
  title={Synthetically Expressive: Evaluating gesture and voice for emotion and empathy in VR and 2D scenarios},
  author={Du, Haoyang and Chhatre, Kiran and Peters, Christopher and Keegan, Brian and McDonnell, Rachel and Ennis, Cathy},
  booktitle={Proceedings of the 25th ACM International Conference on Intelligent Virtual Agents},
  pages={16:1--16:10},
  year={2025}
}

@misc{agrawal2025seamlessinteractiondyadicaudiovisual,
      title={Seamless Interaction: Dyadic Audiovisual Motion Modeling and Large-Scale Dataset}, 
      author={Vasu Agrawal and Akinniyi Akinyemi and Kathryn Alvero and Morteza Behrooz and Julia Buffalini and Fabio Maria Carlucci and Joy Chen and Junming Chen and Zhang Chen and Shiyang Cheng and Praveen Chowdary and Joe Chuang and Antony D'Avirro and Jon Daly and Ning Dong and Mark Duppenthaler and Cynthia Gao and Jeff Girard and Martin Gleize and Sahir Gomez and Hongyu Gong and Srivathsan Govindarajan and Brandon Han and Sen He and Denise Hernandez and Yordan Hristov and Rongjie Huang and Hirofumi Inaguma and Somya Jain and Raj Janardhan and Qingyao Jia and Christopher Klaiber and Dejan Kovachev and Moneish Kumar and Hang Li and Yilei Li and Pavel Litvin and Wei Liu and Guangyao Ma and Jing Ma and Martin Ma and Xutai Ma and Lucas Mantovani and Sagar Miglani and Sreyas Mohan and Louis-Philippe Morency and Evonne Ng and Kam-Woh Ng and Tu Anh Nguyen and Amia Oberai and Benjamin Peloquin and Juan Pino and Jovan Popovic and Omid Poursaeed and Fabian Prada and Alice Rakotoarison and Rakesh Ranjan and Alexander Richard and Christophe Ropers and Safiyyah Saleem and Vasu Sharma and Alex Shcherbyna and Jia Shen and Jie Shen and Anastasis Stathopoulos and Anna Sun and Paden Tomasello and Tuan Tran and Arina Turkatenko and Bo Wan and Chao Wang and Jeff Wang and Mary Williamson and Carleigh Wood and Tao Xiang and Yilin Yang and Julien Yao and Chen Zhang and Jiemin Zhang and Xinyue Zhang and Jason Zheng and Pavlo Zhyzheria and Jan Zikes and Michael Zollhoefer},
      year={2025},
      eprint={2506.22554},
      archivePrefix={arXiv},
      primaryClass={cs.CV},
      url={https://arxiv.org/abs/2506.22554}, 
}

@article{Peebles2022DiT,
  title={Scalable Diffusion Models with Transformers},
  author={William Peebles and Saining Xie},
  year={2022},
  journal={arXiv preprint arXiv:2212.09748},
}
}

\clearpage
\setcounter{page}{1}
\maketitlesupplementary
\appendix

\section{Additional Results}
Since this paper was submitted, we have extended our benchmark with an additional model from the authors of the Seamless Interaction dataset~\cite{agrawal2025seamlessinteractiondyadicaudiovisual}. The Seamless model is a 250M-parameter Diffusion Transformer~\cite{Peebles2022DiT} trained with a conditional flow-matching objective on the Seamless Interaction dataset, which we evaluated in a zero-shot generalisation setting using the BEAT2 test set under the same conditions as before. As part of this extended evaluation, we also collected new votes for previous systems and system pairs, altogether extending our public database of human preference ratings from 16,000 to 20,000 votes.

\begin{figure*}
    \centering
    \includegraphics[width=\linewidth]{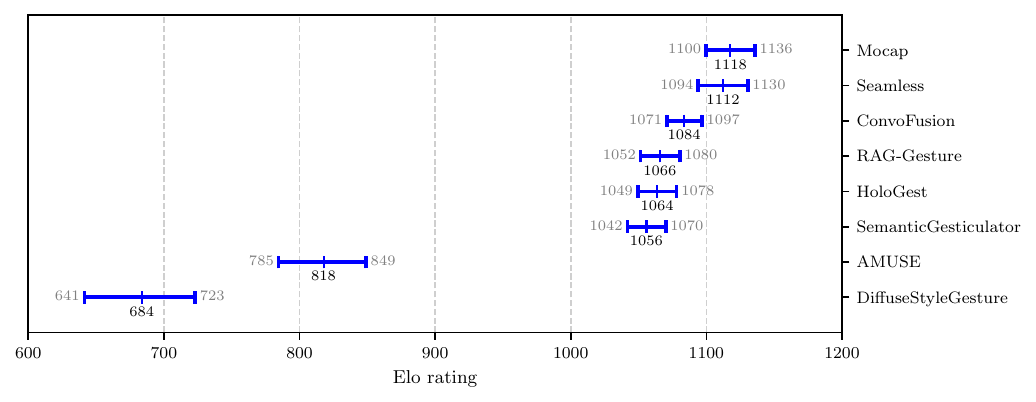}
    \caption{Updated Elo ratings after adding the Seamless model to the benchmark.}
    \label{fig:seamless-elo}
\end{figure*}

\begin{figure}
    \centering
    \includegraphics[width=\linewidth]{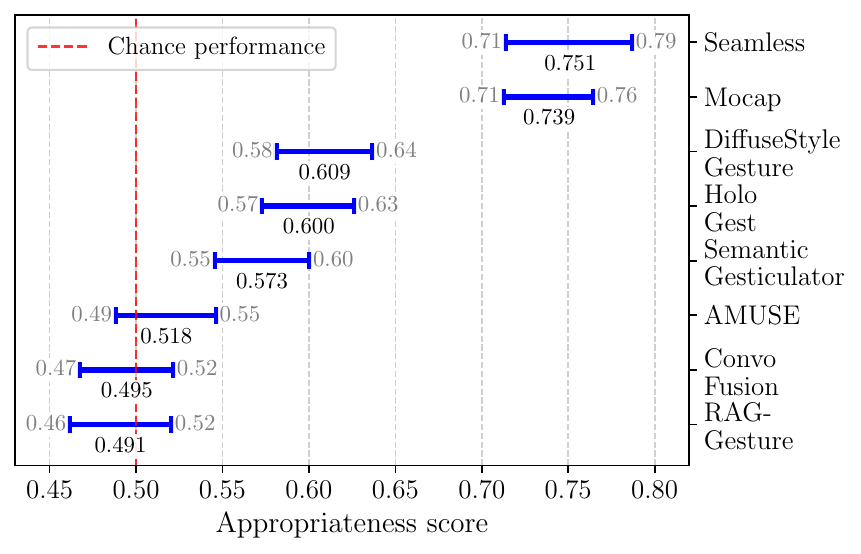}
    \caption{Updated appropriateness scores after adding the Seamless model to the benchmark.}
    \label{fig:seamless-mismatch}
\end{figure}

We present the extended evaluation results in \cref{fig:seamless-elo} and \cref{fig:seamless-mismatch}. To our surprise, the Seamless model not only matched BEAT2 motion capture in terms of motion realism, but also achieved near-indistinguishable appropriateness scores in our audio-mismatching evaluation. This is a breakthrough result that underlines the importance of evolving datasets and standardised evaluations for tracking the state-of-the-art. We emphasise, however, that these high ratings do not imply that the generated motion is perfect: the BEAT2 dataset itself contains motion-capture artifacts, and its actors exhibit varying levels of expressivity and alignment in their gestures. Similarly, our mismatching evaluation is just a first (albeit important) step towards measuring multimodal alignment: there is a strong need for reliable evaluation methods for measuring, e.g., semantic or emotional alignment of gestures.

\section{Additional Details on our Evaluation Protocol} \label{supp:protocol}
\subsection{Test Segment Selection}
\label{sssec:speech-segment-selection}
Following prior large-scale evaluations in speech \cite{perrotin2023blizzard} and gestures \cite{kucherenko2021large,kucherenko2024evaluating,kucherenko2023genea}, we use short speech-gesture segments as the stimuli for our user studies.
Using this one set of speech segments as the basis for all user studies carried out on the BEAT2 dataset, we control for the effect of the speech on system behaviour (e.g., the same speakers are always represented in the same way in every evaluation).

In particular, we curate 108 evaluation segments from the BEAT2 English test set, covering all speakers. The segments are randomly sampled complete sentences, manually filtered for artifacts like flickering and self-intersection.  We selected four segments for most speakers, and eight for Scott and Wayne due to their higher mocap quality. This larger segment count, compared to typical evaluations (e.g., GENEA Challenges) allows for more reliable user studies whilst leaving room for analysis on the stimulus level. The criteria for selecting speech segments for the user studies were as follows:
\begin{itemize}
\item Each segment should correspond to one or more complete sentences.
\item Segment duration should be within the range of 7.0 to 12.0 seconds.
\item Segments should be disjunct (no overlap).
\item Finally, the BEAT2 SMPL-X motion capture for the segments should not contain any major artefacts.
\end{itemize}
The use of complete sentences is more pleasing to test-takers and means that every segment starts and ends at a sentence boundary.
Sentences were identified automatically using the text transcription provided by the dataset.
The specific duration range was chosen based on an informal evaluation by paper authors on all test-set speakers, which indicated that segments shorter than seven seconds too often contained no gesturing at all, whereas segments longer than twelve seconds were difficult to pay sufficiently close attention to throughout, or varied more in quality to the extent that they were more difficult to assign a rating to.

% Only keeping segments where the recorded motion is free of major artefacts is important, since our objective is to compare new gesture-generation methods against the realism of actual human motion, not against the realism achieved by the specific 3D motion-extraction methodology. (The presence of artefacts in the ostensibly natural motion in \citet{yoon2022genea,kucherenko2024evaluating} may explain their counter-intuitive finding that synthetic motion was preferred over natural motion for one of the systems they evaluated.) Concretely, we eliminated (1) all segments containing \emph{flicking}, which we define as instant, physically impossible changes in pose, and most instances of (2) mesh penetration/self-intersection.
Although a lot of the test dataset contains somewhat awkward finger poses due to the difficulty of tracking fingers, this was deemed less visually distracting and would be more drastic to exclude, so it was not considered grounds for exclusion.

The only cases where mesh penetration were permitted were (2a) when the penetration visually resembled clothing or tissue giving way to light pressure, or (2b) where the penetration occurred due to the aforementioned poor finger tracking, as long as the fingers at worst were merely seen clipping into each other, and not passing through each other to the other side at any point.
We decided to retain segments satisfying (2b) for the evaluations because these instances of mesh penetration are associated with poses having gestural importance, such as interwoven hands or finger against palm, which are important not to exclude.

After the segment selection was performed, we measured the potential rhythm bias of our alignment evaluation by comparing our evaluation segments to the semantic labels of BEAT2; we found that ${\approx}59\%$ of segments fully contain at least one semantic gesture, even though we did not intentionally select segments with semantic annotations. In other words, our benchmark heavily oversamples semantic gestures compared to their natural frequency, and is not unnecessarily unfair towards models aiming at semantic generation.

Overall, gesture generation aims for realistic and expressive animation rather than replication of the dataset. This distinction is important due to pose estimation artifacts and the natural variation of human expression, and necessitates careful selection of evaluation segments. Overlooking this step, as most evaluations in our survey do, can lead to lower scores for the reference human motion, and ultimately, imprecise results.

\subsection{\statusWIP{3D Visualisation}}
\label{ssec:visualisation}
%To create the video stimuli shown to the crowdsourced evaluators, we develop a Blender-based rendering pipeline. to be open-sourced upon paper acceptance.
%To avoid retargeting errors, we directly visualise the SMPL-X models \cite{pavlakos2019expressive}, with added textures for increased realism and the face hidden by a mask (as the unrelated facial motion may distract evaluators).
Visualisation plays a critical role in the evaluation of gestural motion. 
Prior work has demonstrated that the quality and type of visualisation can significantly impact the outcome of evaluations \cite{ng2024audio2photoreal}. 
Therefore, it is essential to standardise the visualisation pipeline to ensure that comparisons across systems are made on equal terms. 
In line with the direction of the field towards increasingly photorealistic avatars \cite{ng2024audio2photoreal, chatziagapi2025avflow}, we aim to provide high-quality, realistic visualisation. 
High-quality renderings have been found to aid human raters in distinguishing between better and worse motion, providing clearer, more consistent evaluation results \cite{ng2024audio2photoreal}.

\begin{figure}
    \centering
    \includegraphics[width=\linewidth]{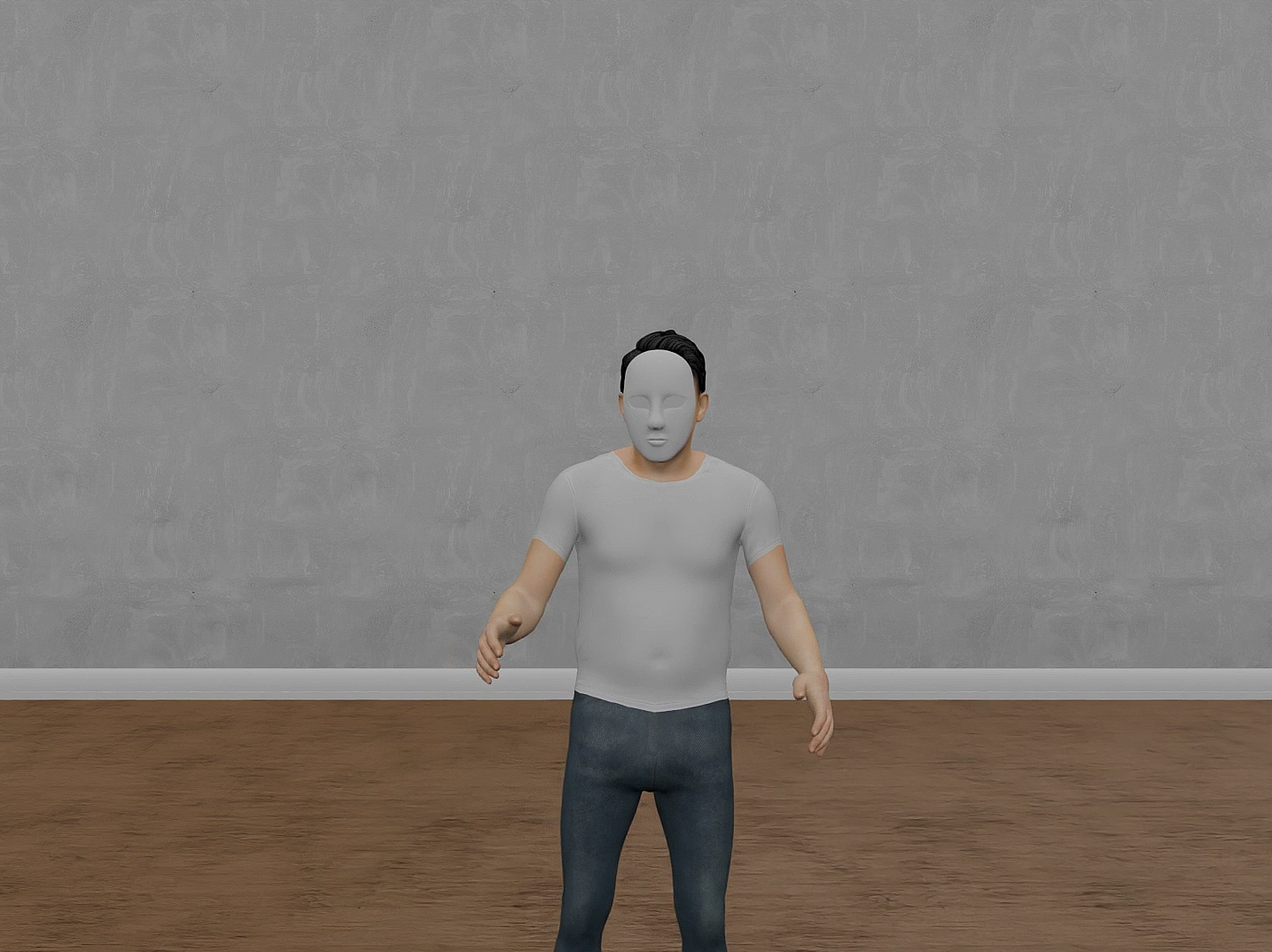}
    \caption{A video frame showing a gesturing SMPL-X avatar (male variant) rendered using the Blender visualiser.}
    \label{fig:blender}
\end{figure}

Following the approach of the GENEA Challenges \cite{yoon2022genea,kucherenko2024evaluating}, our visualisation (\cref{fig:blender}) includes full-body motion with root-node translation and rotation. This captures the character’s positioning and stance with respect to the camera, leading to a more lifelike and expressive visualisation than methods that restrict animation to only the upper body \cite{kucherenko2021large, zhang2024semantic, pang2023bodyformer, chhatre2024amuse}.
    
Since BEAT2 is in the SMPL-X format, we use SMPL-X meshes for our visualisation, which offer an anatomically accurate body shape and proportion that matches each speaker \cite{pavlakos2019cvpr}.
Although SMPL-X includes gendered meshes, we opted to use the gender-neutral mesh for all characters, which is further modified by the per-speaker body shape parameters.
This decision was made because the gendered meshes often resulted in increased self-intersection when visualising motion from the BEAT2 test set.
However, the gender-neutral mesh lacks features such as hair and clothing and does not replicate realistic skin tones or facial characteristics.
The male- and female-presenting speakers are distinguished through different SMPL-X textures: male characters use the default white shirt texture, and female characters use the pink one, matched to the perceived gender of the voice. 
These textures also introduce variation in skin tone. 
To enhance realism, we added a hair prop and applied a displacement map to the clothing, improving its appearance and reducing the flatness seen in the SMPL-X mesh. 
Due to the absence of gaze information and inaccurate lip sync in BEAT2, we covered the face with a white mask. 
This avoids distracting visual artefacts that could negatively influence test-taker perception.

Camera positioning was carefully determined to match the approximate viewpoint of a listener being addressed by the speaker. 
To achieve this, we calculated the mean root-node (hip) translation over the animation frames and used it to normalise the speaker's position in the scene. 
The camera was then statically placed based on this average position to ensure consistent framing across clips.
This setup allows for full visibility of the speaker’s hand and arm gestures, though depending on the magnitude of motion, speakers may appear slightly closer to or further from the camera. In rare cases, if the root translation is particularly large, the character may briefly move out of frame.

Furthermore, we excluded the feet from view, cutting the frame at approximately knee level. 
This was a deliberate choice motivated by known issues with foot-ground interactions, such as sliding and ground penetration, which are both common in synthetic motion and easily detectable by human observers. 
If shown, these artefacts would likely overshadow the gestural qualities under evaluation, as crowdsourced participants tend to focus on the most visually salient errors (see, for instance, the results in \cref{subs:juice_realism}).
By omitting the feet, we help raters concentrate on the gestures themselves, aligning with the primary evaluation objective.

It is also worth noting that while foot-ground contact issues are primarily governed by straightforward laws of physics rules and can potentially be resolved through post-processing, upper-body gesturing is much less dictated by physical laws and is instead rooted in communicative and cultural conventions. 
This arguably makes gesturing a deeper and more challenging problem to solve in the long term, compared to issues of character interaction with the ground plane. For the record, while the human evaluations are conducted based on cropped visualisations, the automatic metrics described in \cref{supp:automatic-evaluation} operate on the full-body pose and motion data.

The rendering environment was kept neutral, using only ambient lighting without shadows.
This setup speeds up rendering while preserving sufficient visual fidelity.  A simple indoor background was chosen to minimize distractions and keep the viewer’s attention on the animated character.

\subsection{User-Study Setup}

\begin{figure*}
    \centering
    \includegraphics[width=0.8\textwidth]{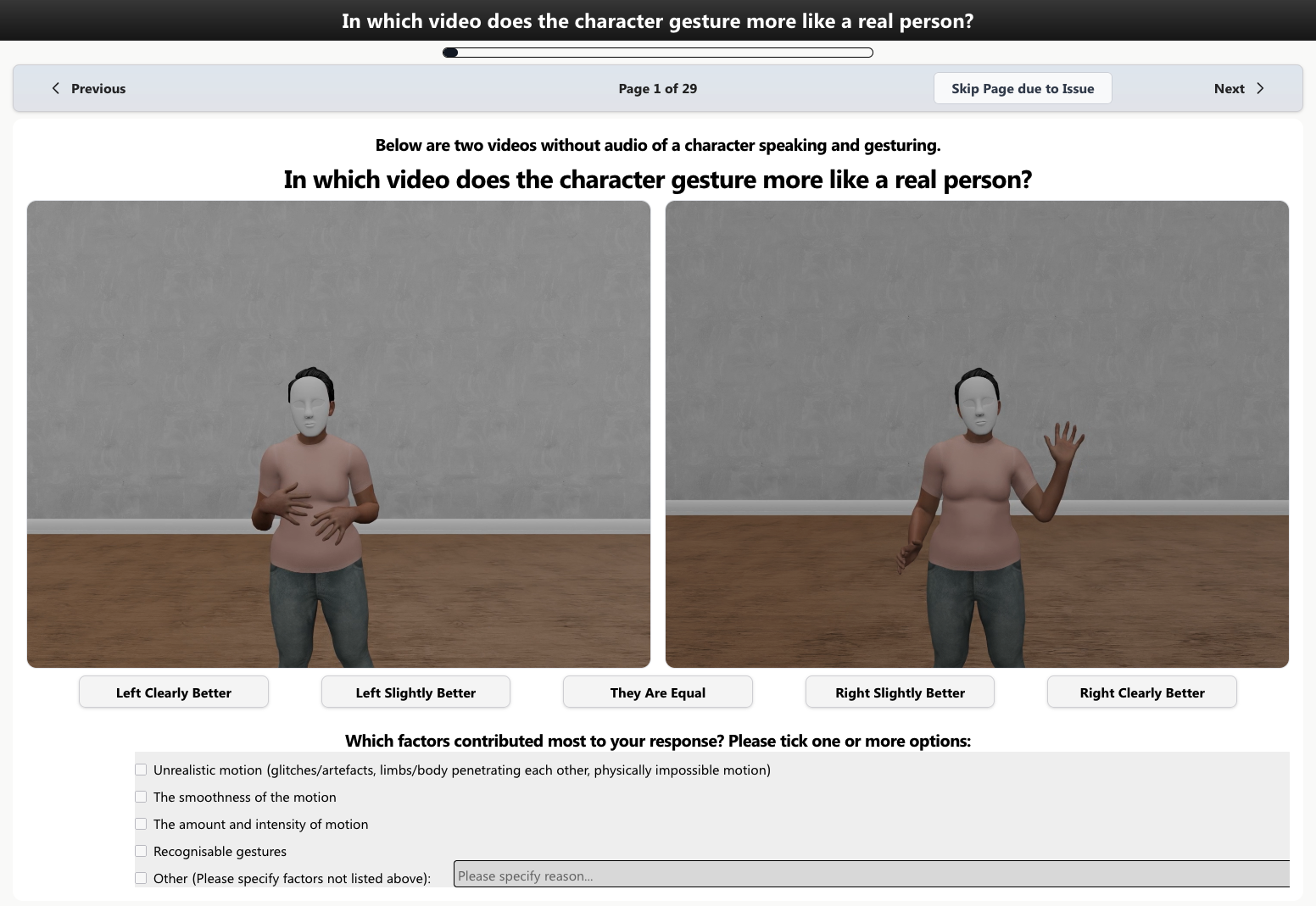}
    \caption{A screenshot of the GUI for the user studies, specifically from a motion-realism test with the current screen containing stimulus videos of the female avatar. The JUICE options are disabled with a grey background since no response to the main question has been selected yet.}
    \label{fig:gui}
\end{figure*}

For the user-study screens, we propose the layout and phrasing (for instructions and response options) shown in \cref{fig:juice_questions}. Our implementation of this interface, used for the experiments in \cref{sec:experiments}, is illustrated in \cref{fig:gui}. We suggest presenting each crowdsourced participant with 25 screens, leading to 25 pairwise votes collected in total (barring technical issues).
\subsubsection{Details on Participant Recruitment}
\label{ssec:study_participants}
As participant recruitment may highly depend between evaluation setups, we do not aim to standardise it in our protocol. Regardless, we share important details from our evaluations in \cref{sec:experiments} below.

Test-takers are recruited through the \href{https://www.prolific.co/}{Prolific} crowdsourcing platform. To be eligible to participate, they are required to reside in any of six English-speaking countries (Australia, Canada, Ireland, New Zealand, the United Kingdom, and the USA) and to have English as their first language. The number of participants and their demographics can be found in \cref{tab:demographics}. No Prolifc user is allowed to participate in the same user study more than once, although this constraint is not enforced between different user studies. Remuneration is set at with 5.25 GBP for a successfully completed test, corresponding to a median of 12.6 GBP hourly rate quoted by the Living Wage Foundation in the UK, computed from the approximate study duration of 25 minutes. 

\subsubsection{Attention Checks}
\label{sssec:attention-checks}
Consistent with best practices in crowdsourced evaluations, our protocol includes attention checks to ensure that test-takers are paying attention to the task.
These take the form of a message ``[Attention check] Please choose `\textit{R}'.'', with \textit{R} being one of the five response options underneath the videos, chosen at random. Four attention checks are inserted into each user study, evenly spaced from the 20\% until the 80\% progress mark. Test-takers that fail any attention check are removed from the statistical analyses; those that fail more than one are rejected without pay. (Prolific's policies do not permit rejecting test takers due to a single failed attention check.)

For the realism evaluation, the attention-check message is presented as high-contrast, easy-to-read text superimposed on one of two otherwise normal video stimuli in a pair. For the speech appropriateness evaluation, each test taker is subjected to two visual (text-based) attention-checks as above, along with two audio attention checks, in which the video is unaffected but the speaker audio in one of the videos is partly replaced by a synthetic voice speaking the same message. In all cases, attention-check messages do not appear until a few seconds into each attention-check video, so that test-takers who only pay attention the first seconds are likely to fail the checks. All test-taker responses given in response to attention checks is excluded from the statistical analyses. Finally, in the case of technical errors such as videos not loading, participants may skip up to three study screens; when a fourth skip occurs, the study is terminated, and a manual review is triggered to establish whether the participant should be paid. This means that each test taker who successfully completes a user study contributes between 23–25 total responses (comprising a one of five possible preference indications and the associated responses to the JUICE questions).

\begin{table}
    \centering
    \small
    \caption{Demographic statistics of crowdsourced test takers that participated in our motion realism evaluation and our mismatching study for speech-gesture appropriation. The age is given as an average and a standard deviation.}
    \label{tab:demographics}
    \begin{tabular}{@{}lccccccccccc@{}}
    \toprule
    \multirow{2}{*}{Study} & \multirow{2}{*}{\Centerstack{Test-\\takers}} & \multicolumn{6}{c}{Country of residence} & \multicolumn{3}{c}{Sex} & \multirow{2}{*}{\Centerstack{Age\\(years)}} \\
    \cmidrule(lr){3-8} \cmidrule(lr){9-11}
            &     & US  & UK  & CA & AUS & IE & NZ & M   & F   & N/A & \\
    \midrule
    Realism & 336 & 202 & \hphantom{0}72  & 48 & 10  & 2  & 2    & 198 & 137 & 1   & $39\pm12$ \\
    Approp. & 311 & 154 & 123 & 19 & \hphantom{0}8   & 3  & 2    & 175 & 132 & 3   & $39\pm13$  \\
    \bottomrule
    \end{tabular}
\end{table}

%Finally, after responding to the 25 video pairs but before submitting the study, participants fill in a short questionnaire to gather broad, anonymous demographic information.

%\begin{summaryItems}
%\item Random selection, 4 segments per speaker except for those with the best capture quality
%\item 100 segments is large compared to many user studies such as the GENEA Challenges, but still few enough that each stimulus will be assessed numerous times in the user studies, so as to allow stimulus-level statistical analysis
%\end{summaryItems}

\section{Additional Details on Our Benchmarking}

\subsection{Statistical Analysis for Motion Realism}
\label{sssec:realism-statistics}
Our statistical analysis is based on the pairwise preference data acquired from the evaluations.
%All responses are first filtered to retain only complete trials where the respondent watched both videos and registered a preference or a tie.
%We also discard any trials where either video fails basic quality checks, such as being corrupted or having mismatched frame rates, to avoid introducing bias.
We standardise the condition labels to canonical forms and convert the raw choices into triplets of the form $(\text{model}_A, \text{model}_B, \text{winner})$, as required by our scoring algorithm.
A ``clear'' preference counts as two wins for the winner, whereas a ``slight'' preference only counts as one; ties count as half a win and half a loss for both models in the presentation.
%Before fitting any models, we consolidate all tie indicators into a single \emph{tie} label and ensure that each system appears at least once as both \textit{model\_A} and \textit{model\_B}.

To transform the pairwise preferences into a continuous ranking, we use the Bradley-Terry Elo-style model advocated by the Chatbot Arena team \cite{chiang2023update}. This approach preserves the interpretability of classical Elo ratings while avoiding the dependence on update order, which can distort results in online systems with large $K$ values. Specifically, we consider the latent skill of each system as a real-valued parameter $e$ and postulate that for any pair $(A,B)$ the log-odds of $A$ beating $B$ are equal to $e_A - e_B$ divided by a scale constant.
Under a logistic link, this assumption yields the Bradley-Terry probability \cite{bradley1952rank}, which is maximized in a single-batch optimization rather than incrementally.
Following generally accepted standards of Elo calculation we set the scale to 400, so that a 200-point difference corresponds to 76\% probability of winning.
This reflects practices in the game of chess, for example.
The model also assumes that the maximum likelihood estimates of the ratings are approximately Gaussian when the number of pairwise comparisons is large, allowing for easy calculation of standard errors. We exploit this asymptotic normality to derive Wald confidence intervals for each rating and to propagate uncertainty when computing derived quantities such as predicted win rates.
Because the pair frequencies are unbalanced, we additionally perform non-parametric bootstrapping over the original trials to guard against violations of the Gaussian approximation. In each bootstrap replicate, we sample battles with replacement, fit the Bradley-Terry model, and record the resulting set of Elo ratings.
All optimization is done using \texttt{scikit-learn}'s unconstrained logistic regression solver, which reliably converges to our dataset within seconds.

\begin{figure*}[!t]
\begin{boxedminipage}[H]{\textwidth}
\centering
Below are two videos without audio of a character speaking and gesturing.\\

\vspace{\baselineskip}
\fbox{Left video\vphantom{Right}}\quad\fbox{Right video}\\

\vspace{\baselineskip}
In which video does the character gesture more like a real person?

\fbox{Left clearly better} \fbox{Left slightly better} \fbox{They are equal} \fbox{Right slightly better} \fbox{Right clearly better}\\
%\fbox{\Longstack[c]{Left is\\clearly better}}\fbox{\Longstack[c]{Left is\\slightly better}}\fbox{\Longstack[c]{They are\\equal}}\fbox{\Longstack[c]{Right is\\slightly better}}\fbox{\Longstack[c]{Right is\\clearly better}}

\vspace{\baselineskip}
Which factors contributed most to your response? Please tick one or more options:
\begin{itemize}[label={$\square$}]
\item Unrealistic motion (glitches/artefacts, limbs/body penetrating each other, physically impossible motion)
\item The smoothness of the motion
\item The amount and intensity of motion
\item Recognisable gestures
\item Other (Please specify factors not listed above): \rule{5cm}{0.15mm}
\end{itemize}
\end{boxedminipage}\\
\vspace{\baselineskip}
\begin{boxedminipage}[H]{\textwidth}
\centering
Below are two videos of a character speaking and gesturing. Both videos have the same motion, but different speech.\\

\vspace{\baselineskip}
\fbox{Left video\vphantom{Right}}\quad\fbox{Right video}\\

\vspace{\baselineskip}
In which video do the character’s movements fit the speech better?

\fbox{Left clearly better} \fbox{Left slightly better} \fbox{They are equal} \fbox{Right slightly better} \fbox{Right clearly better}\\

\vspace{\baselineskip}
Which factors contributed most to your response? Please tick one or more options:
\begin{itemize}[label={$\square$}]
\item Fit the rhythm and timing of the speech better
\item Emphasised the correct part (or parts) of the speech
\item Better matched the content and meaning of the speech
\item Better fit for the emotion of the speech
\item Other (Please specify factors not listed above): \rule{5cm}{0.15mm}
\end{itemize}
\end{boxedminipage}
\caption{Questions and response options in the two types of user studies, also showing their schematic layout in the user-study GUI. For a screenshot of the GUI see \cref{fig:gui}.}
\label{fig:juice_questions}
\end{figure*}%

\subsection{Statistical Analysis of Speech Appropriateness}
\label{sssec:appropriateness-statistics}
For the statistical analysis, we use the basically same setup as described in \cref{sssec:realism-statistics} for motion realism:
clear preference responses count double compared to slight preferences, and ties (``They are equal'') count as half a win and half a loss.
The only difference is that wins for the matched stimulus are assigned the value 1 and wins for the mismatched stimulus are assigned a 0.
The resulting average \emph{appropriateness score} (essentially a modified win rate) is then a number between zero and one.

The rest of the analysis is the same as for motion realism.
We use the exact same test-taker-level bootstrap methodology to obtain confidence intervals, based on quantiles of the bootstrap distribution.
%To test if differences in appropriateness score between two conditions are statistically significantly different from zero, we likewise look at the differences between the appropriateness scores of any two systems in the bootstrap samples. We compute two-sided $p$-values regarding the score difference in the same way as for motion realism and apply Benjamini-Hochberg FDR correction \cite{benjamini2000adaptive} afterwards.

\subsection{JUICE Scores for Motion Realism}
\label{subs:juice_realism}

\begin{figure}
    \centering
    \includegraphics[width=\linewidth]{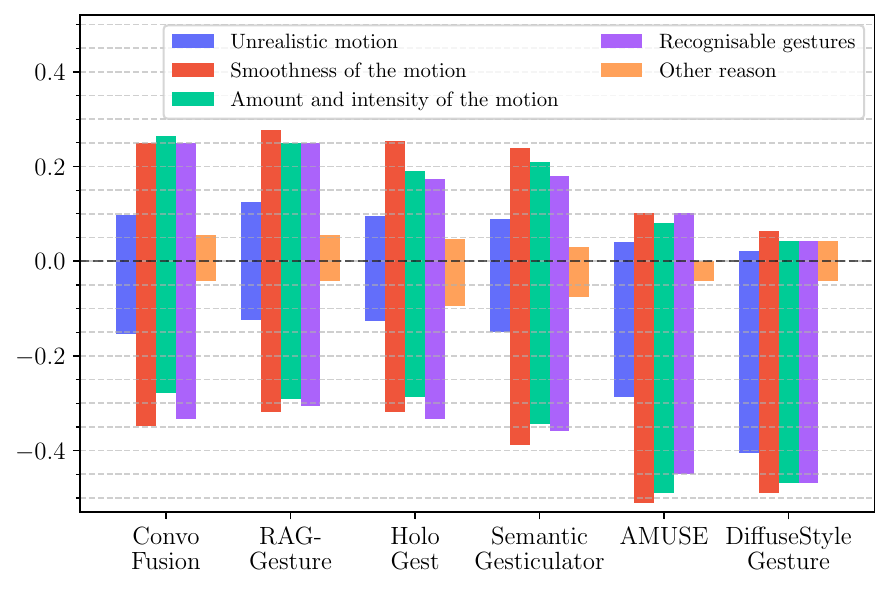}
    \caption{Frequency of JUICE options chosen for each model during the motion-realism evaluation in comparisons to the motion-capture condition, ignoring ties. Bar plots above zero show frequency among winning outcomes; bar plots below zero correspond to frequencies among losing outcomes, both relative to the total number of non-tie comparisons for the given model.}
    \label{fig:juice-realism-result}
\end{figure}

Although we collect JUICE responses for all presentations where there was not a tie, we here focus on analysing the JUICE responses when comparing each of the six initial synthetic systems to the mocap topline.
The distribution of these is graphed as a bar chart in \cref{fig:juice-realism-result}.
(All JUICE responses, including free text for the ``Other'' option, are featured in the data we release.)

\cref{fig:juice-realism-result} was created by, for each system, first counting how often each of the five JUICE options (tickboxes) were ticked when that system was pitted against the BEAT2 mocap.
(Ties, i.e., ``They are equal'' are ignored since they did not generate JUICE responses.)
%shows the distribution of normalised JUICE responses, obtained when comparing each of the six generative systems to the BEAT2 reference motion capture.
%First, we counted how often each of the five JUICE reasons were selected for a given system when paired with the BEAT2 mocap, irrespective of whether the system won or lost the comparison (ties, i.e., ``They are equal'', did not produce JUICE responses).
After that, we normalised these values into percentages, where 100\% would mean that the specific JUICE option in question was ticked every single one of these presentations.
Finally, we split the percentages by whether or not they were associated with a win (the bar pointing up from zero) or a loss (the same bar extending down below zero instead) for the system in question.
%so that the reason sums to one within each model.
Our normalisation brings forth the qualitative profile of a model by compensating for imbalances in the total number of responses, which vary because comparisons with large perceptual differences pause early.
Furthermore, strong wins and losses were counted as one win or loss instead of two in our analyses of JUICE responses in this paper.
Together, this setup means that the percentages in the plot are biassed towards factors that correspond to subtle differences.

Across the top four systems, the \emph{Smoothness of the motion}, the \emph{Amount and intensity of the motion}, and \emph{Recognisable gestures} were each ticked at comparable rates.
This broad similarity mirrors the tight Elo clustering observed earlier and suggests that evaluators focus on nuanced aspects of kinematics when the realism gap to the motion capture is small.
The catch-all category \emph{Other reason} was used relatively sparingly for every system, suggesting that the four predefined options captured most salient perceptual differences.
Analysis of the free-text responses is left as future work.
%with little residual ambiguity.

The most notable deviation from the general uniformity of response rates to pre-defined JUICE options is the frequency with which the \emph{Unrealistic motion} option was chosen.
Although selected less often, it is disproportionately associated with AMUSE and especially DiffuseStyleGesture, the two systems that occupy the lower end of the Elo ratings in this study.
This clear pattern supports an interpretation that visible artefacts, such as jerks, implausible limb trajectories, and temporal discontinuities, are a primary cause of dispreference when present and must be not be generated for competitive performance.

Past gesture-synthesis systems have been criticised for producing ``marginally natural gestures that appear more like well-timed hand waving, are not communicative and have little meaning'' \cite{nyatsanga_comprehensive_2023}.
As such, it might a-priori be expected that synthetic systems may struggle to produce distinctive and recognisable communicative gestures, e.g., iconic and metaphoric gestures.
We therefore find it surprising to see that ``recognisable gestures'' did not show any apparent advantage for mocap over synthetic gestures.
Although it is possible that strong contemporary systems have improved on the issues pointed out by \citet{nyatsanga_comprehensive_2023}, e.g., with the RAG-Gesture system \citep{mughal2024raggesture} employing retrieval-augmented generation, it is also possible that this option might need to be replaced by another formulation and/or be complemented by additional instructions in the future that more clearly communicate its intention to crowdsourced test takers.

\subsection{JUICE Scores for Speech-Gesture Appropriateness}
\label{subs:juice_appropriateness}

\begin{figure}
    \centering
    \includegraphics[width=\linewidth]{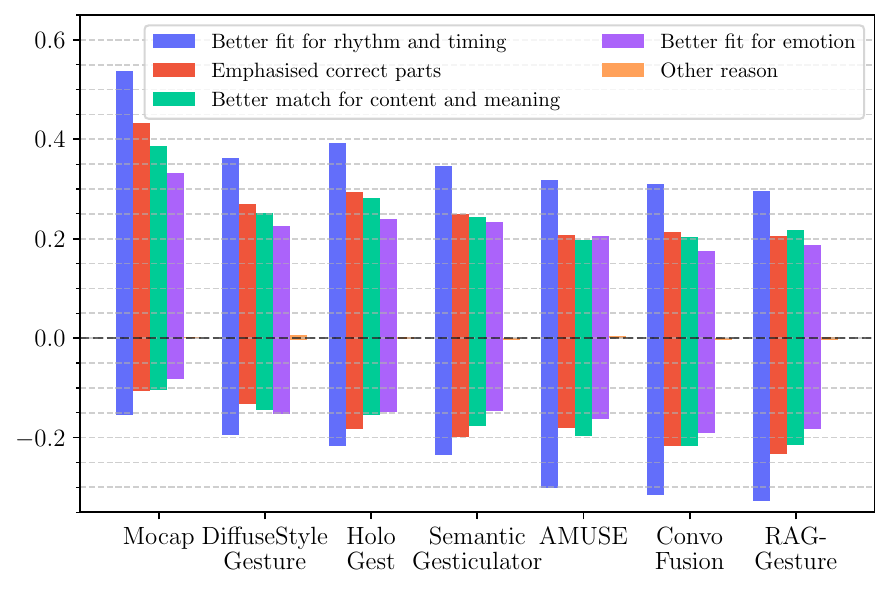}
    \caption{Frequency of JUICE responses chosen in the speech-gesture appropriateness study, for each model, when compared against its mismatched counterpart. Positive values are the frequency among winning outcomes; negative values correspond to the frequency among losing outcomes, both relative to the total number of non-tie comparisons.}
    \label{fig:juice-appropriateness-results}
\end{figure}

The distribution of normalised JUICE responses for gesture–speech appropriateness is shown in \cref{fig:juice-appropriateness-results}. Like for the realism analysis, the number of responses for each option were first accumulated for every condition, regardless of the outcome of the trial. Then, the numbers were converted to proportions that sum to one within each condition.
However, whereas the earlier \cref{fig:juice-realism-result} specifically graphed the data from cases where test takers compared clips from each artificial system to BEAT2, the appropriateness study never asks test takers to compare conditions directly, but only to assess matched and mismatched video clips within each  condition.
For this reason, the Mocap condition is included in \cref{fig:juice-appropriateness-results} but not in the earlier \cref{fig:juice-realism-result}.

Across all conditions, the reason \emph{Fit the rhythm and timing of the speech} was selected disproportionally often, both when a model was preferred and when it was dispreferred, indicating that temporal alignment is most salient feature for test-taker decisions.
This makes sense, given that the speech segments evaluated were not selected to contain rich semantic grounding or strong emotional colouring, making it so that rhythm naturally becomes the primary distinguishing factor. (See also \citet{saund2021importance}.)
%That said, it is never in a majority.
%That said, it was never ticked even close to half of the time, meaning that, in most cases, one or more of the other boxes was ticked.

The other JUICE options \emph{Emphasised the correct part of the speech}, \emph{Better matched the content and meaning of the speech}, and \emph{Better fit for the emotion of the speech}, were each chosen at similar rates and significantly less often than rhythm, but were still used an appreciable faction of the time.
This implies that participants considered these aspects overall less consistently important for their choice.
The catch-all category \emph{Other reason} was used even more rarely than in the motion-realism JUICE response data, indicating that the predefined options well capture the most important sources of preference in the appropriateness domain.
%consistent with the moderate appropriateness scores reported in \cref{fig:mismatching-results}.

Unlike the \emph{Unrealistic motion} option in \cref{fig:juice-realism-result}, there are no strong indications that certain JUICE options are selected disproportionally often for certain systems, including, say, for RAG-Gesture and Semantic Gesticulator, both of which specifically target improved semantic consistency in their work.
Appropriateness evaluations with segments selected to contain semantic gesturing, or to mismatch between emotions, might alter this balance and could be interesting future work.

\section{Details on Systems Evaluated}
\label{sec:system_details}
In this section, we describe the notable features of each of the six gesture-generation systems evaluated in the main paper, as well as the adaptation steps performed by each model's original authors when preparing their submissions.

\subsection{DiffuseStyleGesture \cite{yang2023diffusestylegesture}}
\label{sssec:diffusestylegesture}

DiffuseStyleGesture aims to generates high-quality, speech-synchronized 3D co-speech gestures through a diffusion model architecture. The generation process ensures outputs exhibit robust temporal audio-gesture synchronization and stable kinematics. A notable feature is the incorporation of seed gestures for initialisation, providing control over the generation process, leading to varied and contextually relevant motion.

DiffuseStyleGesture was originally trained on the ZEGGS dataset \cite{ghorbani2023zeroeggs}. To prepare the submission, the authors of the model processed input motion into a comprehensive per-joint feature set \cite{yang2023diffusestylegesture+}. Additionally, a key modification to the published model is the use of a data filtering strategy, exclusively utilising data from a select cohort of professional actors (Actors 2, 3, 4, 7, 10, 15, 16, 17, 18, 21, and 27) to learn from high-fidelity motion exemplars. Generated joint-based positional output is converted to the SMPL-X format by mapping joint rotations (from Euler angles) to SMPL-X axis-angle pose parameters via a predefined joint map, alongside extracting and transforming the root joint's translation and coordinate system for SMPL-X alignment.

This manual conversion from positional data to SMPL-X, adopted to maintain input feature parity with \cite{yang2023diffusestylegesture, yang2023diffusestylegesture+}, may compromise visual fidelity compared to direct SMPL-X feature utilisation in training and generation \cite{liu2024emage}. Therefore, additional post-processing was employed in the form of a minor scaling applied to the root joint's motion to enhance visual stability and mitigate potential drift during front-facing camera evaluation, and a subtle inverse kinematics (IK) adjustment from the feet to the root, which serves as a minor refinement with negligible visual impact on foot placement.

\subsection{Semantic Gesticulator \cite{zhang2024semantic}}
\label{sssec:semanticgesticulator}
Semantic Gesticulator aims to generate high-quality, semantically meaningful gesture animations from speech by combining rhythmic precision with contextual understanding. Unlike prior models that rely solely on direct audio-to-motion mappings, this model introduces a discrete latent motion space via a residual VQ-VAE, enabling compact and diverse motion representations. It uniquely integrates a GPT-based gesture generator with a large language model (LLM)-driven semantic retrieval system, which selects appropriate gestures based on transcript context. A semantics-aware alignment module then fuses rhythmic and semantic information, resulting in gestures that are both expressive and contextually appropriate.

The model's authors adapted the original system by removing the semantic gesture retrieval component and relying solely on the base RVQ+GPT pipeline for audio-to-gesture generation. This simplification allows evaluating the core generative capacity of the model. Additionally, the data preprocessing module was modified to support the SMPL-X representation used in the BEAT2 dataset, ensuring compatibility with our motion format. During training, the RVQ module was configured with a codebook size of 1024 and 4 quantization layers to accommodate the longer and more complex motion sequences present in the full BEAT2-English dataset. The GPT-based gesture generator was trained using the same architecture and settings as described in the original system. No additional postprocessing was applied to the motion output, enabling an unbiased assessment of the model’s raw generation quality.

\subsection{ConvoFusion \cite{mughal2024convofusion}}
\label{sssec:convofusion}
ConvoFusion is a diffusion-based framework for speech- and text-driven gesture synthesis. %that also enables control over both textual and auditory modalities in speech.
%The base framework takes speech and its transcription as input to generate natural-looking gestures.
%The latent diffusion architecture comprises two components:
%(a) a scale-aware temporal VAE that models different body parts separately and represents sequential motion frames using temporally ordered latents, and
%(b) a transformer decoder for diffusion that contains separate cross-attention heads for conditioning on different modalities i.e. speech, gesture and speaker identity.  
It features a latent diffusion architecture with two components: 1) a scale-aware temporal VAE that models different body parts separately and represents sequential motion frames using temporally ordered latents, and 2) a transformer decoder for diffusion that contains separate cross-attention heads for conditioning on different modalities i.e. speech, gesture and speaker identity.
\par
Originally, this framework was not trained on BEAT2 and is therefore modified to accommodate the SMPL-X input representation. The scale-aware VAEs are trained independently for four body parts: upper body, hands, face, and lower body. Following this, the base latent diffusion framework is adapted to the updated VAEs.
% which entails that latent diffusion framework generates motion in SMPL-X format. 
Additionally, the speech representation is upgraded from mel-spectrograms to wav2Vec embeddings~\cite{baevski2020wav2vec}. Leveraging the temporal structure of VAE latents, the framework is capable of auto-regressively generating long-form motion in time-windowed chunks. To perform auto-regressive rollout, it first generates the initial 10 seconds of motion, then uses the last 1 second of that output as seed motion to generate the next 9 seconds. This seed motion maintains continuity across steps through diffusion-based outpainting. At each step, overlapping motion segments are linearly blended with the previous ones, resulting in a single coherent motion sequence.

\subsection{RAG-Gesture \cite{mughal2024raggesture}}
\label{sssec:rag-gesture}
RAG-Gesture aims to generate not only natural looking but also semantically meaningful gestures. It achieves this by first training a base latent-diffusion framework for co-speech gesture generation (similar to ConvoFusion~\cite{mughal2024convofusion}), and then leveraging retrieval augmented generation during inference to inject semantically meaningful exemplars. The generated gestures are therefore sampled from the base distribution of a diffusion model, while also being semantically grounded in explicit domain knowledge, like gesture types or discourse relations. The method is agnostic to the choice of retrieval algorithm; in the original paper, two approaches were presented: one based on an LLM's understanding of gesture type, and the other grounded in discourse-based linguistic analysis of the speech.

% Since the original system is already trained on the BEAT2 dataset in the SMPL-X format, ... and follows its input representation.
% Specifically it generates hand, body, face motion along with the translation of the character from a single model. 
% To generate long-form motions, it performs autoregressive generation in the chunks of 10 second motions. 
% Specifically, it generates first 10 seconds and then uses the last 1 second of the output as a seed motion to generate next 9 seconds in the next step.
% Seed motion keeps the motion consistent across steps by the use of diffusion outpainting.
% At the end, overlapping motion segments are linearly blended, forming one complete motion sequence for given speech.
The system is inherently trained on BEAT2 dataset and follows its input representation, therefore no adaptation to the trained model is made. Specifically it generates hand, body, face motion along with the translation of the character from a single model. As the framework follows the temporal VAE structure, it also performs long-form motion generation in chunks of 10-second time windows through autoregressive rollout (\cref{sssec:convofusion}). Consequently, retrieval algorithm is not used for the overlapping motion frames and RAG is performed for the newly generated motion. For evaluation, LLM-driven Gesture Type algorithm is used for RAG.

\subsection{AMUSE \cite{chhatre2024amuse}}
\label{sssec:amuse}
AMUSE is an emotional, speech-driven model for 3D body animation. It converts audio filter-bank features into three disentangled latent vectors that separately encode (1) linguistic content, (2) emotional state, and (3) speaker style. The speech encoder is a Vision Transformer (ViT) \cite{pmlr-v139-touvron21a} adapted to operate on filter-bank images. These vectors condition a latent-diffusion model~\cite{rombach2022high} that generates gesture motion sequences. After training, AMUSE can synthesize 3D human gestures directly from speech while allowing users to combine content, emotion, and style, for example, pairing the content vector of a source speech with the emotion and style vectors from a different one. Stochastic sampling of the diffusion noise term yields diverse gesture variants that preserve the chosen emotional expressivity.

AMUSE was developed on the BEAT2 SMPL-X data, with the same dataset splits that we use. However, there are two differences between the submission format and the data processing of the original model that require adaptation.
First, AMUSE puts emphasis on upper-body gesticulation rather than locomotion, therefore it discards the eight lower-body joints of the SMPL-X body. This was resolved by augmenting the model outputs with static lower-body joints. Second, AMUSE can only generate 10-second motion sequences, while the submission system normally expects a single, coherent motion sequence for each test-set file. As a workaround, the AMUSE submission contains 7--12 second motion clips, corresponding to the full set of speech segments described in \cref{sssec:speech-segment-selection}. Clips shorter than 10 seconds were generated by padding the audio input with silence, and discarding surplus motion frames from the output. Clips longer than 10 seconds were artificially created by blending two clips, $c_1$ and $c_2$, using spherical linear interpolation (SLERP), where $c_1$ is generated on the first 10 seconds of the segment, and $c_2$ is generated from the last two seconds of $c_1$ and the remaining portion of the segment.

\subsection{HoloGest \cite{cheng2025hologest}} \label{sssec:hologest}
HoloGest aims to generate physically plausible and vivid co-speech gestures by addressing limitations in current diffusion-based methods, which often use a single noise distribution for full-body gestures despite their differing characteristics. It tackles this issue by decoupling body parts to learn separate noise distributions and introduces motion priors to enhance physical plausibility, effectively reducing unnatural phenomena like jitter and sliding. Additionally, HoloGest employs an adversarial generation approach to accelerate the denoising process, requiring only 50 steps (0.7 seconds) to produce 2 seconds of gestures, making it suitable for real-time performance. These strategies enable HoloGest to deliver highly realistic and dynamic gestures while maintaining computational efficiency.
\par
The published version of HoloGest features two motion priors trained on external datasets: one for the finger motion, and another for the root trajectory. In contrast, the HoloGest submission ensures fairness towards other participating systems by removing the finger prior, and retraining the trajectory prior on the BEAT2 training set, without relying on external datasets. Furthermore, the independent diffusion generation channel for facial expressions was removed, therefore the submission only contains body- and hand motion.
%Considering the fairness of the ranking list, this system has removed the publicly available datasets used for motion priors in the original paper. All the priors are now trained solely on the BEAT2 training set. %In addition, unlike what was described in the original paper, this system has also removed the independent diffusion generation channel for facial expressions. Currently, this system have only decomposed the holistic co-speech gestures into channels for the upper and lower body along with both hands. Regarding the motion priors, this system have removed the part of the finger priors and have not cited the external SignAvatars dataset either. At the same time, for the trajectory prior, this system has made a rotation operation on the original BEAT2 data. Specifically, this system has rotated the root joint by 90 degrees along the X-axis to ensure that it is in an upright standing position when visualize in Blender4.3.2 (the original data was lying flat). After that, all the generation results are directly generated by the model without any further post-processing.

\section{Experiments on Automatic Metrics}
\label{supp:automatic-evaluation}

We provide evaluation results using a curated set of automatic metrics, often called objective metrics, primarily selected based on their frequent use in recent gesture-generation research. While human evaluation ultimately determines overall performance, automatic metrics may serve as a complementary tool to benchmark and analyse system behaviour efficiently and at scale. 
% We report and analyse the metric results across different systems and investigate how well current automatic measures reflect human ratings.

We report results of seven metrics. \textbf{Fréchet Gesture Distance (FGD)} measures the Fréchet Distance between human motion and generated motion distributions on a learnt feature space \cite{yoon2020speech, liu2024emage}. \textbf{Fréchet Distance on Geometric and Kinetic Features ($\text{FD}_g$ and $\text{FD}_k$)} \cite{alexanderson2023listen, ng2024audio2photoreal}; $\text{FD}_g$ measures the Fréchet Distance between the distributions of static pose data from human and generated motion. $\text{FD}_k$, on the other hand, compares the distributions of inter-frame pose differences (i.e., motion velocity). \textbf{Beat Alignment (BA)} evaluates the alignment between the beats in the input speech and those in the generated motion \cite{li2021ai,liu2022learning}. 
%Originally proposed in the domain of dance motion generation \cite{li2021ai}, this metric has recently been adopted in gesture generation as well \cite{liu2024emage, mughal2024convofusion, zhi2023livelyspeaker, chen2024language}. It is also referred to as Beat consistency or Beat constancy in some studies. 
%Note that we used mapping from audio beats to motion beats for BA computation. 
\textbf{Semantic Relevance Gesture Recall (SRGR)} compares human motion and generated motion by evaluating the proportion of correctly recalled joints only over segments containing semantic gestures \cite{liu2022beat}. \textbf{Pose Diversity ($\text{DIV}_{pose}$)} evaluates how diverse the generated poses are within each motion sequence by computing the average deviation of individual poses from the mean pose. \textbf{Sample Diversity ($\text{DIV}_{sample}$)} measures the diversity across multiple generated motion samples for the same input, indicating the stochastic variability of the model’s outputs.

%\subsection{\statusDone{Results and Discussion}}
%We report the results of automatic metric evaluations conducted on the test set of the BEAT2 dataset. The evaluation includes human motion data and outputs from generative systems. It is important to note that all metrics ignore root-node position and translation, and are therefore insensitive to visual artefacts like foot sliding and the character floating off the ground or facing the wrong way.

\sisetup{detect-weight, mode=text, detect-family}
\newcommand{\bestsym}[1]{\bfseries #1}

\begin{table*}[!ht]
    \centering
    \caption{Automated evaluation of gesture generation models using a set of objective metrics. Human motion capture data is included for reference. The best value for each metric among systems is \textbf{boldfaced}. For BA and $\text{DIV}_\text{pose}$ metrics, values closer to the human reference motion are considered better.}
    \label{tab:automatic_results}
    \begin{tabular}{@{}l lSSSSSSS@{}}
    \toprule
        & Condition & \text{FGD} $\downarrow$ & $\text{FD}_\text{g}$ $\downarrow$ & $\text{FD}_\text{k}$ $\downarrow$ & $\text{BA}$ $\rightarrow$ & SRGR $\uparrow$ & $\text{DIV}_\text{pose}$ $\rightarrow$ & $\text{DIV}_\text{sample}$ $\uparrow$ \\
    \midrule
    & Motion capture & 0.000 & 0.000 & 0.000 & 0.645 & 1.000 & 8.302 & \text{--} \\
    \midrule
    \multirow{7}{*}[0pt]{\centering\rotatebox{90}{\footnotesize\textbf{Systems}}}
    % & EMAGE & 0.361 & 4.528 & 0.051 & 0.604 & 0.305 & 11.722 & 0.000 \\
    & HoloGest & 0.625 & 0.972 & 0.059 & 0.539 & \bestsym{0.469} & \bestsym{7.733} & 0.011 \\
    & RAG-Gesture & 0.515 & \bestsym{0.660} & \bestsym{0.035} & \bestsym{0.648} & 0.427 & 10.092 & 0.013 \\
    & DiffuseStyleGesture & 7.110 & 10.128 & 0.099 & 0.608 & 0.312 & 9.598 & 0.001 \\
    & Semantic Gesticulator & \bestsym{0.473} & 0.749 & 0.043 & 0.681 & 0.398 & 10.993 & \bestsym{0.020} \\
    & ConvoFusion & 0.600 & 0.817 & 0.040 & 0.611 & 0.448 & 8.911 & 0.013 \\
    & AMUSE* & 0.785 & 0.997 & 0.041 & 0.757 & 0.394 & 9.552 & 0.018 \\
    \bottomrule
    \multicolumn{9}{l}{\scriptsize*AMUSE results are affected by motion discontinuities stemming from its lack of long sequence support.} \\
    \end{tabular}
\end{table*}

Table \ref{tab:automatic_results} presents the results on the test set of the BEAT2 dataset. RAG-Gesture shows strong performance for distribution-based motion-quality metrics (FGD, $\text{FD}_g$, $\text{FD}_k$), as well as BA. HoloGest shows the best SRGR and $\text{DIV}_{\text{pose}}$. Semantic Gesticulator yields the best FGD and richest run-to-run variability ($\text{DIV}_{\text{sample}}$).
Note that the FGD values are not directly comparable to those in \citet{liu2024emage} due to differing data sizes (see \citet{chong2020effectively}): we used all audio in the test set and all five random samples submitted for each system to obtain as many data points as possible for better distribution fitting.

We examined the correlation between the results of the automatic metrics and the subjective human ratings. First, we looked at the motion-realism-related metrics, FGD, $\text{FD}_g$, and $\text{FD}_k$. According to the user study, the ConvoFusion, RAG-Gesture, HoloGest, and Semantic Gesticulator systems achieved relatively high Elo ratings, while AMUSE and DiffuseStyleGesture had lower ratings compared to the other systems. When roughly dividing the systems into these two groups, we observed that the high Elo-rating group consistently outperformed the low Elo-rating group on the FGD and $\text{FD}_g$ metrics, which aligns with the user ratings.

Next, regarding speech-gesture appropriateness, we considered the BA and SRGR metrics. Here, we found a substantial discrepancy between the automatic metrics and human ratings. For example, RAG-Gesture achieved the best BA score, but had the lowest user rating in the user study. Similarly, although HoloGest, RAG-Gesture, and ConvoFusion achieved high SRGR scores, they did not demonstrate clear appropriateness in the human evaluations.

Inspired by the GENEA Challenges \cite{kucherenko2024evaluating}, we also conducted a quantitative analysis of the correlation between automatic metric scores and subjective human ratings. Given the limited number of systems, the correlation analysis serves as a reference and should not be interpreted as providing strong evidence or conclusive findings. Specifically, we computed the correlation between Elo ratings (representing human preferences) and each automatic metric using Kendall’s $\tau$ rank correlation \cite{kendall1948rank}. For human ratings on motion realism, all motion quality metrics exhibited moderate negative correlations (between $-0.4$ and $-0.6$, consistent with the findings for FGD in \citet{kucherenko2024evaluating}), while SRGR -- despite being more closely related to speech-gesture alignment -- showed the highest positive correlation (0.73). For speech-motion alignment, BA demonstrated a moderate correlation (0.5), whereas SRGR showed virtually no correlation ($-0.14$). However, none of these correlations were statistically significant ($p < 0.05$).
% This kind of analysis will be much more meaningful once the leaderboard has been running for some time and features data on more systems.

Overall, our findings highlight that whilst automatic metrics can provide useful insights and facilitate early evaluation, they remain insufficient to replace human evaluation in gesture generation. The discrepancies -- especially in speech–gesture alignment -- underscore the limitations of current objective measures and the continued necessity of human evaluation. 
% These results emphasise the necessity of further research to develop more robust and validated objective metrics, as well as the need for good human evaluation for the foreseeable future.

\clearpage

\end{document}